\pdfoutput=1

\documentclass[11pt]{article}

\usepackage[final]{ACL2025}

\usepackage{times}
\usepackage{latexsym}

\usepackage{amsmath}
\usepackage{multirow}
\usepackage{bbm}
\usepackage{graphicx}
\usepackage{amssymb}
\usepackage{booktabs}
\usepackage{lscape}
\usepackage{pifont}
\usepackage{algorithm}
\usepackage{caption}
\usepackage{subcaption}
\usepackage{CJKutf8}
\usepackage{makecell}
\usepackage{xcolor}
\usepackage{svg}
\usepackage{todonotes}
\newcommand\benchmarkname{\textcolor{black}{\textsc{WorldCuisines}}}

\newcommand\vqaname{\textcolor{black}{\textsc{WC-VQA}}}

\newcommand\kbname{\textcolor{black}{\textsc{WC-KB}}}

\newcommand{\Hquad}{\hspace{0.37em}} 

\usepackage[T1]{fontenc}

\usepackage[utf8]{inputenc}

\usepackage{microtype}

\usepackage{inconsolata}

\title{$\benchmarkname$: A Massive-Scale Benchmark for Multilingual and Multicultural Visual Question Answering on Global Cuisines}

\author{
Genta Indra Winata\thanks{$\text{ }$ These authors contributed equally. This is an open-source project, and the work was done outside of their affiliations. Contacts: \url{genta.winata@capitalone.com} and \url{frederikus.hudi.fe7@naist.ac.jp}.}$\hspace{1.7mm}$$^{\spadesuit,1,2}$, 
Frederikus Hudi$^{*\spadesuit,2,3}$, 
Patrick Amadeus Irawan$^{*\spadesuit,2,4}$, \\
\textbf{
David Anugraha$^{*\spadesuit,5}$, Rifki Afina Putri$^{*\spadesuit,2,6}$, Yutong Wang$^{\spadesuit,7}$, Adam Nohejl$^{\spadesuit,3}$, 
}\\\textbf{
Ubaidillah Ariq Prathama$^{\spadesuit,4}$, Nedjma Ousidhoum$^{\spadesuit,8}$, Afifa Amriani$^{9}$, 
}\\\textbf{
Anar Rzayev$^{6}$,  Anirban Das$^{1}$, Ashmari Pramodya$^{3}$, Aulia Adila$^{7}$, Bryan Wilie$^{10}$, 
}\\\textbf{
Candy Olivia Mawalim$^{7}$, Ching Lam Cheng$^{11}$, Daud Abolade$^{12,13}$, 
}\\\textbf{
Emmanuele Chersoni$^{14}$, Enrico Santus$^{9}$, Fariz Ikhwantri$^{9}$, Garry Kuwanto$^{15}$, 
}\\\textbf{
Hanyang Zhao$^{16}$, Haryo Akbarianto Wibowo$^{17}$, Holy Lovenia$^{2}$, 
}\\\textbf{
Jan Christian Blaise Cruz$^{2,17}$, Jan Wira Gotama Putra$^{9}$, Junho Myung$^{6}$, 
}\\\textbf{
Lucky Susanto$^{18}$, Maria Angelica Riera Machin$^{3}$, Marina Zhukova$^{19}$, 
}\\\textbf{
Michael Anugraha$^{9}$, Muhammad Farid Adilazuarda$^{2,17}$, Natasha Santosa$^{20}$, 
}\\\textbf{
Peerat Limkonchotiwat$^{2,21}$, Raj Dabre$^{22}$, Rio Alexander Audino$^{4}$, 
}\\\textbf{
Samuel Cahyawijaya$^{2,23}$, Shi-Xiong Zhang$^{1}$, Stephanie Yulia Salim$^{7}$, Yi Zhou$^{8}$, 
}\\\textbf{
Yinxuan Gui$^{11}$, David Ifeoluwa Adelani$^{\clubsuit,12,24,25,26}$, En-Shiun Annie Lee$^{\clubsuit,5,27}$,
}\\\textbf{
Shogo Okada$^{\clubsuit,7}$, Ayu Purwarianti$^{\clubsuit,2,4}$, Alham Fikri Aji$^{\clubsuit,2,17,18}$, Taro Watanabe$^{\clubsuit,3}$, 
}\\\textbf{
Derry Tanti Wijaya$^{\clubsuit,15,18}$, Alice Oh$^{\clubsuit,6}$, Chong-Wah Ngo$^{\clubsuit,11}$,
}
\\
$^{1}$Capital One$\Hquad$ $^{2}$SEACrowd$\Hquad$ $^{3}$NAIST$\Hquad$ $^{4}$ITB$\Hquad$ $^{5}$UofT$\Hquad$ $^{6}$KAIST$\Hquad$ $^{7}$JAIST$\Hquad$ $^{8}$Cardiff University
\\
$^{9}$Independent$\Hquad$ $^{10}$HKUST$\Hquad$ $^{11}$SMU$\Hquad$ $^{12}$Masakhane$\Hquad$ $^{13}$University of Lagos$\Hquad$ $^{14}$HK PolyU$\Hquad$
\\
$^{15}$Boston University $^{16}$Columbia University $^{17}$MBZUAI $^{18}$Monash University$ $ $^{19}$UCSB
\\
$^{20}$Tokyo Tech $^{21}$AI Singapore $^{22}$NICT $^{23}$Cohere $^{24}$McGill $^{25}$MILA \\ $^{26}$Canada CIFAR AI Chair 
$^{27}$Ontario Tech
\\
$\Hquad^\spadesuit$Main Authors $\Hquad$ $^\clubsuit$Senior Authors
}

\begin{document}
\maketitle

\begin{abstract}
Vision Language Models (VLMs) often struggle with culture-specific knowledge, particularly in languages other than English and in underrepresented cultural contexts. To evaluate their understanding of such knowledge, we introduce $\benchmarkname$, a massive-scale benchmark for multilingual and multicultural, visually grounded language understanding. This benchmark includes a visual question answering (VQA) dataset with text-image pairs across 30 languages and dialects, spanning 9 language families and featuring over \emph{1 million data points}, making it the largest multicultural VQA benchmark to date. It includes tasks for identifying dish names and their origins. We provide evaluation datasets in two sizes (12k and 60k instances) alongside a training dataset (1 million instances). Our findings show that while VLMs perform better with correct location context, they struggle with adversarial contexts and predicting specific regional cuisines and languages. To support future research, we release a knowledge base with annotated food entries and images along with the VQA data.
\end{abstract}

\section{Introduction}
\begin{figure}[!t]
    \centering
    \includegraphics[width=\linewidth]{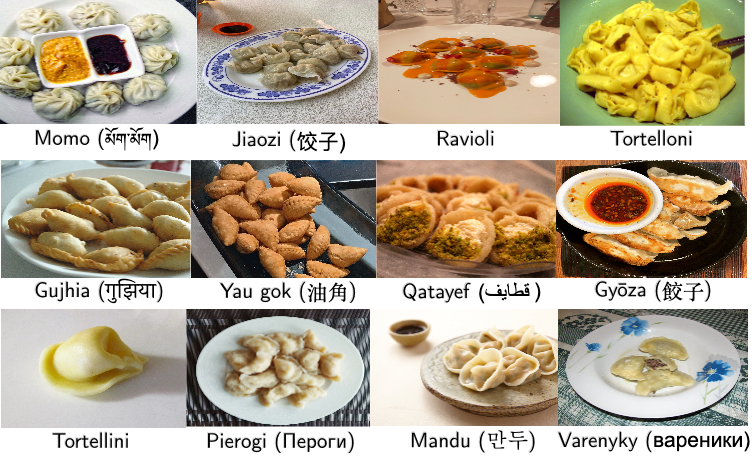} 
    \caption{Images of stuffed pasta and dumplings from our dataset showcase a similar culinary concept across different cultures: wrapping meat, dairy (such as cheese), or vegetables in dough. These dishes can be prepared in various ways, including pan-frying, deep-frying, steaming, or boiling.}
    \label{fig:stuffed_pasta_dumpling_example}
    \vspace{-2mm}
\end{figure}
\begin{table*}[!t]
\centering
\resizebox{\textwidth}{!}{
    \begin{tabular}{lrcccccc}
    \toprule 
    & \textbf{\# VQA} & \textbf{\# Lang./Dialect$^\dagger$} & \textbf{\# Countries} & \textbf{\# Food Entries} & \textbf{\# Images} & \textbf{Parallel Data} & \textbf{License} \\ \midrule
    FoodieQA~\cite{li2024foodieqa} & 659 & 2 & 1 & 60 & 389 & $\times$ & CC BY-NC-ND 4.0 \\ 
    World Wide Dishes~\cite{magomere2024you} & 765 & 131 & 63 & 765 & 301 & $\times$ & CC-BY 4.0 \\
    xGQA~\cite{pfeiffer2022xgqa} & 12,578 & 8 & 8 & N/A & 398 & $\checkmark$ & CC-BY 4.0 \\
    MaXM$^\ddagger$~\cite{changpinyo2023maxm} & 2,142 & 7 & 7 & N/A & 335 & $\times$ & Custom \\
    EVJVQA~\cite{nguyen2023vlsp2022} & 33,790 & 3 & 1 & N/A & 4,909 & $\times$ & N/A \\
    CulturalVQA~\cite{nayak2024benchmarking} & 2,378 & 1 & 11 & N/A & 2,328 & $\times$ & N/A \\
    SEA-VQA~\cite{urailertprasert2024sea} & 1,999 & 1 & 8 & N/A & 515 & $\times$ & Custom \\
    CVQA~\cite{romero2024cvqa} & 9,000 & 26 & 28 & 1,834 & 4,560 & $\checkmark$ & Various\\
    IndiFoodVQA~\cite{agarwal2024indifoodvqa} & 16,716 & 1 & 1 & 255 & 414 & $\times$ & N/A \\
    \midrule
    $\vqaname$ & \textbf{1,152,000} & 30 & 189 & 2,414 & 6,045 & $\checkmark$ & CC BY-SA 4.0\\ 
    \bottomrule
    \end{tabular}
}
\caption{Data statistics for $\vqaname$ compared to existing VQA datasets. The data samples are sourced from their respective publications. $^\ddagger$The reported numbers are based on their human-annotated test set. $^\dagger$This entry includes the language variations we collected for all languages.}
\label{tab:data-statistics}
\vspace{-3.5mm}
\end{table*}

Food is an essential medium for the exchange of regional cultures, serving to connect diverse peoples and traditions~\cite{wahlqvist2007regional}. Analyzing various culinary practices provides valuable insights into the cultural values, historical narratives, and social customs of the communities that produce and consume these foods~\cite{holtzman2006food}. Furthermore, food plays a significant role in shaping language, which serves as a proxy for cultural knowledge~\cite{freedman2021food}. Food choices often reflect intricate community histories, societal transformations, and both individual and collective identities, thereby creating a rich tapestry of cultural expression~\cite{almerico2014food}. The relationship between culture and food is dynamic; both evolve in tandem over time, resulting in the emergence of new dishes that are influenced by historical culinary traditions~\cite{anderson2014everyone}.

As a result, similar food concepts can be found across different countries, reflecting a shared human culinary heritage. Researchers use food as a proxy to model and analyze cultural dynamics, helping to quantify cultural differences across regions~\cite{adilazuarda2024towards}. Many cultures have developed their own versions of ``stuffed pasta'' or ``dumplings'', each with unique ingredients and preparation methods, often known by different names~\cite{gallani2015dumplings}, as illustrated in Figure~\ref{fig:stuffed_pasta_dumpling_example}. Small details like how the dumpling is shaped can signal the cultural background. Conversely, some dishes share the same name but have different meanings; for example, ``jelly'' in the U.S. refers to a fruit spread, while in the U.K. and parts of Asia, it refers to a gelatinous dessert~\cite{poppe1992gelatin,abe2013tokyo}. This culinary diversity presents a challenge for Vision Language Models (VLMs), which must accurately recognize and differentiate food items based on cultural context for applications like food recognition. These models navigate the complexities of names, ingredients, and preparation styles that vary widely across regions. VLMs have shown effectiveness in text captioning~\cite{liu2024improved,liu2024visual} and have been adapted to support multiple languages~\cite{geigle2023mblip,shin2024x}.

However, there is limited research on evaluating the multicultural capabilities of VLMs, particularly in terms of multilinguality. The study by~\citet{romero2024cvqa} introduce visual question answering (VQA) from a multicultural perspective, but it mainly focuses on knowledge and situational context at a specific moment, which does not fully assess the ability of VLMs to reason and differentiate between cultures within a single question. Moreover, another study on food VQA is limited to Chinese culture and does not explore the broader spectrum of global cultures~\cite{li2024foodieqa}. An earlier investigation into cultural bias in language models also found that cultural knowledge is lacking~\cite{naous2023having}. Therefore, further research is necessary to address these limitations and enhance our understanding of VLMs' multicultural and multilingual capabilities.

\begin{figure*}[!th]
    \centering
    \includegraphics[width=\linewidth]{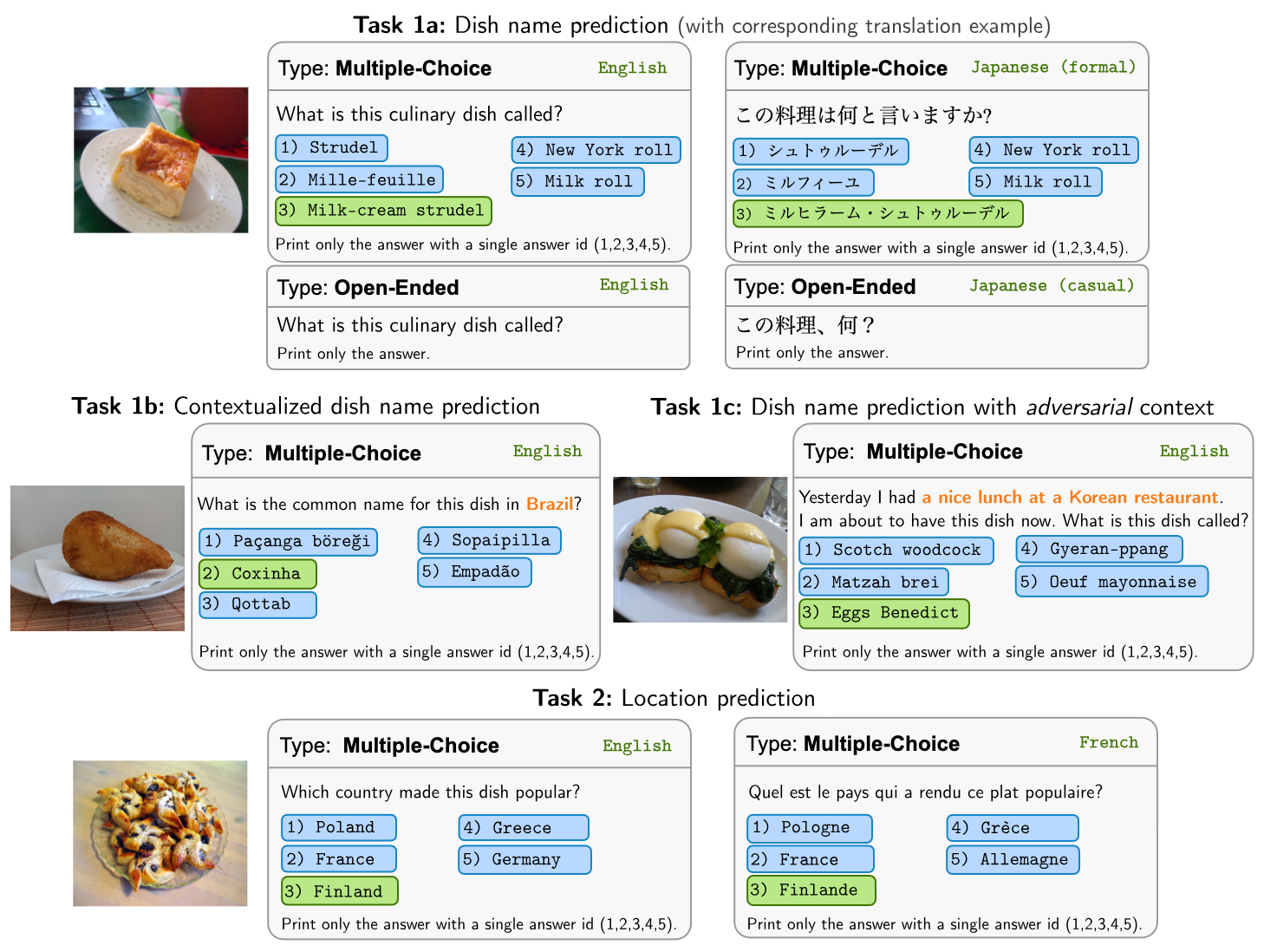}
    \caption{$\vqaname$ in $\benchmarkname$ comprises two primary tasks: (1) predicting dish names and (2) predicting regional cuisines. Task 1 is further divided into three subtasks: (a) no-context, (b) contextualized, and (c) adversarial. We also include two answer types: multiple-choice question (MCQ) and open-ended question (OEQ).}
    \label{fig:tasks}
    \vspace{-1.5mm}
\end{figure*}

To facilitate a comprehensive analysis of multilingual and multicultural research, we develop resources for evaluating VLMs. Table~\ref{tab:data-statistics} summarizes how our work compares to previous studies. Our benchmark stands out for its cultural diversity, offering more VQA datasets and broader language and dialect coverage. Our major contributions can be summarized in three-fold:
\begin{itemize}
    \item We present $\benchmarkname$, the first massive scale benchmark consisting of 1 million high-quality multilingual and multicultural text-image pairs annotated by native speakers in their local languages.
    We publicly release our resources, i.e., datasets,
    \footnote{We release $\vqaname$ at~\url{https://huggingface.co/datasets/worldcuisines/vqa} and $\kbname$ consisting food, location, cuisine, and prompt templates at~\url{https://huggingface.co/worldcuisines}.} code,\footnote{We release our code at~\url{https://github.com/worldcuisines/worldcuisines}.} and leaderboard\footnote{We release our leaderboard at~\url{https://huggingface.co/spaces/worldcuisines/worldcuisines}.} 
    to advance future research in this rapidly evolving field.
    \item We evaluate open-source and commercial VLMs for cultural awareness through two VQA tasks: predicting dish names from images and context, and identifying their geographical origin. We also assess the impact of context, including adversarial scenarios.
    \item We create multilingual templates for queries and context (such as the questions in QA pairs) while preserving language varieties, including dialects and registers. This is achieved by creating translations that incorporate different inflections, articles, and contractions. Our goal is to ensure naturalness in each translation and to use appropriate inflections for place names.
\end{itemize}

\begin{figure*}[!t]
    \centering\includegraphics[width=0.95\linewidth]{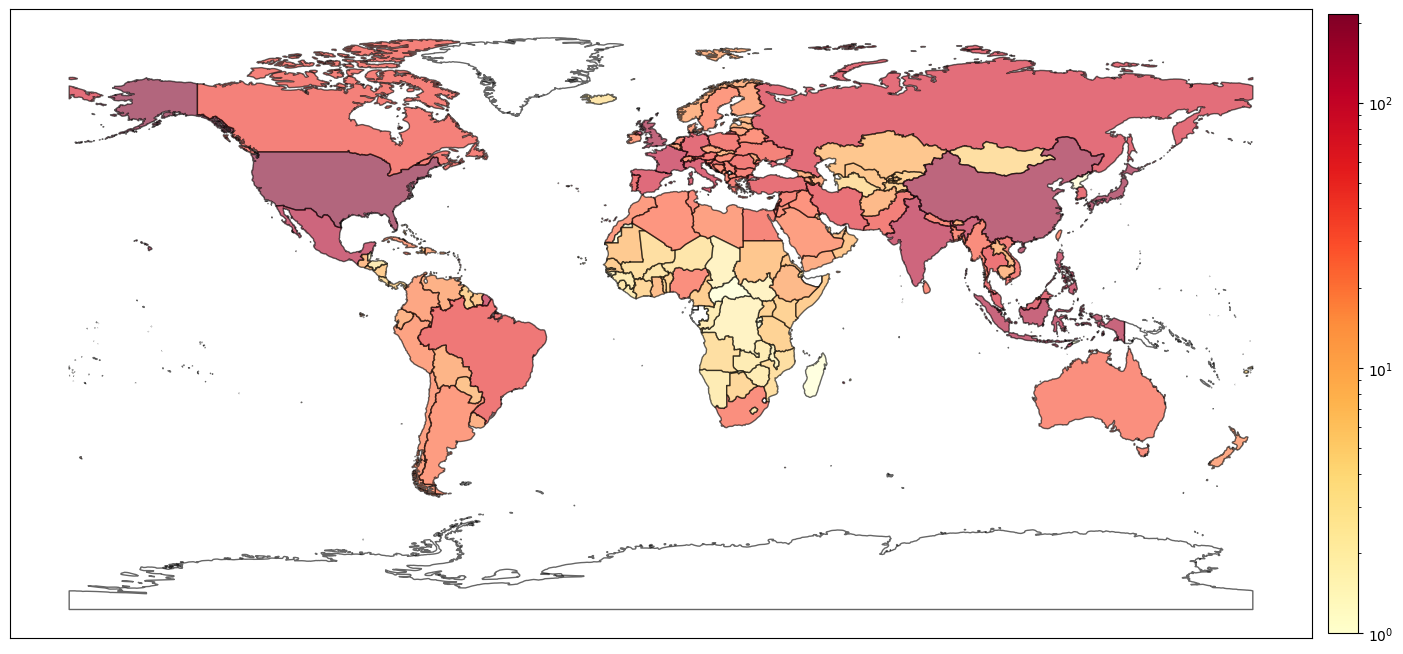}
    \caption{$\benchmarkname$ distribution of food entries by country in the World Map. The food entries are distributed across 189 countries, with the highest concentration found in Asia, Europe, and North America. There are also some entries from the continents of Africa, Oceania, and Central and South America.}
    \label{fig:dish-geodist}
\end{figure*}
\begin{figure*}[!th]
    \centering
    \begin{subfigure}[t]{\textwidth}
        \centering
        \includegraphics[width=\linewidth]{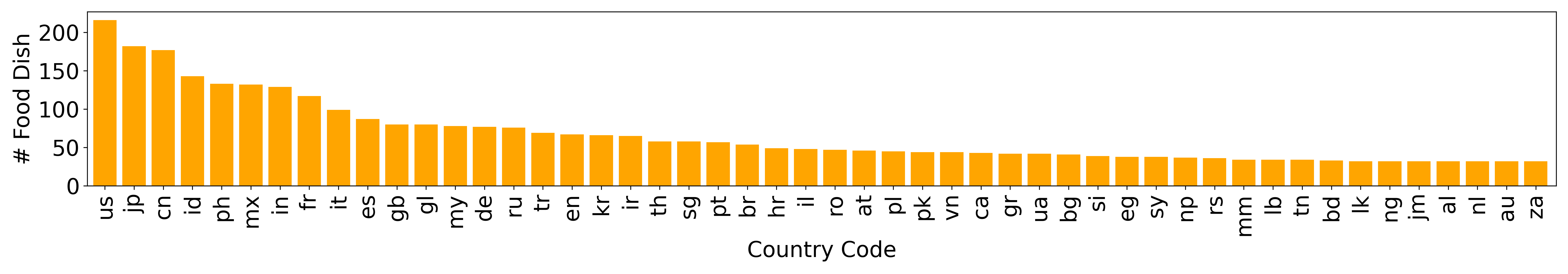} 
        \label{fig:bottom}
        \vspace{-30mm}
    \end{subfigure}
    \caption{Countries by number of assigned dishes, showing the top 50 countries.}
    \label{fig:country_food_dish}
    \vspace{-3mm}
\end{figure*}

\section{$\benchmarkname$}
We propose $\benchmarkname$, an open-source benchmark designed to evaluate the cultural relevance and understanding of VLMs. Figure~\ref{fig:tasks} displays VQA examples in English, alongside selected parallel translations in Japanese and French.

\subsection{Overview}
We develop both a VQA dataset ($\vqaname$) and a curated KB for world cuisines ($\kbname$). The $\vqaname$ dataset is constructed using $\kbname$, which serves as the primary data source. We design two tasks as follows:
\begin{itemize}
    \item \textbf{Task 1:} Dish Name Prediction. This task involves predicting the name of a dish based on its image, a question, and contextual information. It comprises three subtasks, each with distinct query types: \emph{(a) no-context} question, \emph{(b) contextualized} question, and \emph{(c) adversarial contextualized} question.
    \item \textbf{Task 2:} Location prediction. The task is to predict location where the food is commonly consumed and originated given the dish image, question, and a context.
\end{itemize}

\paragraph{$\kbname$.} 
A KB encompassing 2,414 dishes worldwide includes 6,045 images and metadata, covering both coarse-grained (e.g., stew) and fine-grained categories (e.g., beef stew), locations, and regional cuisines. It also features multilingual translations of 90 crowd-sourced prompt templates and 401 parallel data entries (i.e., multilingual information) for location and regional cuisine information.
\vspace{-1mm}
\paragraph{$\vqaname$.} A multilingual parallel VQA dataset with 1 million samples encompassing over 30 languages and dialects, including various varieties and registers, such as formal and casual styles, with high-quality human annotations. The VQA is designed to evaluate models' ability to understand cultural food names and their origins.

\subsection{$\kbname$ Construction}
Our data sources are gathered from Wikipedia\footnote{Wikipedia web pages can be accessed at~\url{https://wikipedia.org}.} and Wikimedia Commons\footnote{Wikimedia Commons web pages can be accessed at~\url{https://commons.wikimedia.org}.} to ensure they can be easily redistributed under an accepted open-source license. The data construction process involves four key steps: (1) dish selection, (2) metadata annotation, (3) quality assurance, and (4) data compilation. Figure \ref{fig:dish-geodist} provides statistics on the regions covered in our dataset, with detailed information available in Table \ref{tab:region_dist} in the Appendix. Figure \ref{fig:country_food_dish} shows the distribution of dish frequencies, highlighting the top 50 countries with the most dishes.

\subsubsection{Dish Selection}
We compile a comprehensive list of dish names sourced from Wikipedia. We manually review pages that feature lists of dishes to determine whether each dish is a specialty unique to a specific culture, as we aim to focus on dishes that have distinct cultural significance. We exclude generic categories, such as ice cream, which lacks a specific cultural association. We ensure that each dish on our list has its own dedicated Wikipedia page. If a dish does not have a Wikipedia page, it is also excluded from our compilation. This meticulous approach ensures that our dataset is both culturally relevant and well-documented.

\begin{table*}[t]
    \centering
    \resizebox{\textwidth}{!}{
    \begin{tabular}{lrrrrrr|rr|r}
        \toprule
        \multicolumn{1}{c}{\multirow{3}{*}{\textbf{\makecell{Data\\Split}}}} & \multicolumn{6}{c|}{\multirow{1}{*}{\textbf{Task 1 (Dish Name)}}} & \multicolumn{2}{c|}{\multirow{1}{*}{\textbf{Task 2}}} & \multicolumn{1}{c}{\multirow{3}{*}{\textbf{\makecell{Total\\\# VQA}}}} \\
         & \multicolumn{2}{c}{\textbf{(a) no-context}} & \multicolumn{2}{c}{\textbf{(b) contextualized}} & \multicolumn{2}{c|}{\textbf{(c) adversarial}} & \multicolumn{2}{c|}{\textbf{(Location)}} & \\
        & \# VQA & \# Images & \# VQA & \# Images & \# VQA & \# Images & \# VQA & \# Images\\
        \midrule
        Train (1M) &  270,300  &  3,383  &  267,930  &  3,555  &  271,770  &  3,589  &  270,000  &  3,361 & 1,080,000 \\
        Test Small (12k) &  3,000  &  100  &  3,000  &  100  &  3,000  &  100  &  3,000  &  100 & 12,000 \\
        Test Large (60k) &  15,000  &  500  &  15,000  &  500  &  15,000  &  499 &  15,000  &  499 & 60,000 \\
        \bottomrule
    \end{tabular}
    }
    \caption{Dataset statistics for $\vqaname$ tasks for train, test small, and test large data splits. Total \#VQA represents the total number of VQA from Task 1 and Task 2.}
    \label{tab:vqa_splits}
\end{table*}

\subsubsection{Metadata Annotation}
Given a dish name and its corresponding Wikipedia page link, we then ask annotators to manually compile metadata based on the provided information. This metadata includes:
\begin{itemize}
    \item \textbf{Visual Representation}: Images sources from Wikimedia Commons are included, along with their license information.
    \item \textbf{Categorization}: Dishes are classified into both coarse-grained (e.g., rice, bread) and fine-grained (e.g., fried rice, flatbread) categories.
    \item \textbf{Description}: Annotators provide a description of each dish based on the content from its Wikipedia page, avoiding the use of the dish's name, origin, or any distinctive keywords that uniquely identify the dish.
    \item \textbf{Cuisine}: The dish's origin cuisine and any cuisines with which it is strongly associated.
    \item \textbf{Geographic Distribution}: This includes the dish's associated countries, area (city or region), and broader continental region.
\end{itemize}
The metadata description, along with the example, is further elaborated in the Appendix Table~\ref{tab:attributes}.

\begin{table*}[!th]
\centering
\resizebox{\textwidth}{!}{
    \begin{tabular}{lrrrrrr|rr|rr}
    \toprule
    \multicolumn{1}{c}{\multirow{3}{*}{\textbf{Model}}} & \multicolumn{6}{c|}{\textbf{Task 1 (Dish Name)}} & \multicolumn{2}{c|}{\multirow{2}{*}{\textbf{\makecell{Task 2 \\ (Location)}}}} & \multicolumn{2}{c}{\multirow{2}{*}{\textbf{Average}}} \\ 
    & \multicolumn{2}{c}{\textbf{(a) no-context}} & \multicolumn{2}{c}{\textbf{(b) contextualized}} & \multicolumn{2}{c|}{\textbf{(c) adversarial}} & & \\
    & \hspace{.8em}MCQ & \multicolumn{1}{c}{OEQ}  & \hspace{2em}MCQ & \multicolumn{1}{c}{OEQ} & \hspace{.8em}MCQ & \multicolumn{1}{c|}{OEQ} & \hspace{.8em}MCQ & \multicolumn{1}{c|}{OEQ} & \hspace{.5em}MCQ & \multicolumn{1}{c}{OEQ} \\ \midrule
    \textbf{Open-Source} & & & & & & & & & & \\
    $\quad$Llava1.6 Vicuna 7B & 34.57 & 1.59 & 43.48 & 4.03 & 34.84 & 1.41 & 32.24 & 9.29 & 36.28 & 4.08 \\
    $\quad$Llava1.6 Vicuna 13B & 40.17 & 2.79 & 48.17 & 5.85 & 39.05 & 2.57 & 37.79 & 10.16 & 41.30 & 5.34 \\
    $\quad$Qwen2 VL Instruct 2B & 41.65 & 7.98 & 42.29 & 8.13 & 39.69 & 6.74 & 47.85 & 14.55 & 42.87 & 9.35 \\
    $\quad$Qwen2 VL Instruct 7B & 61.48 & 6.76 & 67.85 & 10.36 & 53.52 & 6.12 & 55.90 & 21.03 & 59.69 & 11.07 \\
    $\quad$Qwen2 VL Instruct 72B & 74.19 & 12.67 & 80.79 & 21.31 & 62.43 & 8.37 & 61.90 & 27.27 & 69.83 & 17.40 \\
    $\quad$Llama 3.2 Instruct 11B & 59.93 & \underline{18.75} & 64.12 & 22.96 & 53.17 & \underline{13.39} & 57.93 & \underline{31.58} & 58.79 & \underline{21.67} \\
    $\quad$Llama 3.2 Instruct 90B  & \underline{77.69} & 16.93 & \underline{82.92} & \underline{23.60} & 63.96 & 10.87 & \underline{67.87} & 31.31 & 73.11 & 20.68 \\
    $\quad$Molmo-E 1B & 18.81 & 0.01 & 24.22 & 0.23 & 19.55 & 0.01 & 18.97 & 1.54 & 20.39 & 0.45 \\
    $\quad$Molmo-D 7B & 46.01 & 2.89 & 55.95 & 3.66 & 41.61 & 2.31 & 33.35 & 11.45 & 44.23 & 5.08 \\
    $\quad$Molmo-O 7B & 39.96 & 5.15 & 44.93 & 6.03 & 38.41 & 3.51 & 29.81 & 10.07 & 38.28 & 6.19 \\
    $\quad$Pangea 7B$^\ddagger$ & 52.35 & 1.52 & 63.07 & 2.73 & 49.17 & 1.57 & 48.71 & 20.15 & 53.33 & 6.49 \\
    $\quad$Aria 25B & 58.61 & 4.99 & 69.29 & 9.17 & 52.82 & 3.39 & 42.82 & 16.20 & 55.89 & 8.44 \\
    $\quad$Phi-3.5 Vision 4B & 43.37 & 2.91 & 48.71 & 4.23 & 40.87 & 2.07 & 35.01 & 9.22 & 41.99 & 4.61 \\
    $\quad$Pixtral 12B & 56.65 & 1.22 & 70.69 & 2.94 & 52.12 & 1.09 & 46.67 & 14.43 & 56.53 & 4.92 \\
    $\quad$NVLM-D 72B & 69.82 & 4.71 & 78.93 & 10.29 & 52.12 & 2.89 & 51.97 & 16.68 & 63.21 & 8.64 \\
    \midrule
    \textbf{Proprietary} & & & & & & & & & & \\
    $\quad$GPT-4o & \textbf{88.45} & \textbf{21.88} & \textbf{91.57} & \textbf{37.51} & \textbf{82.29} & \textbf{14.79} & 66.52 & \textbf{37.13} & \textbf{82.21} & \textbf{27.83} \\
    $\quad$GPT-4o Mini & 72.80 & 10.28 & 81.65 & 20.87 & 57.76 & 5.72 & 52.37 & 25.79 & 66.14 & 15.66 \\
    $\quad$Gemini 1.5 Flash & 77.05 & 12.81 & 80.97 & 15.16 & \underline{69.13} & 6.46 & \textbf{71.53} & 30.03 & \underline{74.67} & 16.12 \\

    \bottomrule
    \end{tabular}
}
\caption{Accuracy (\%) results of $\vqaname$ for Test Large (60k). MCQ and OEQ indicate multiple-choice question and open-ended question, respectively. Best and second-best are \textbf{bolded} and \underline{underlined}, respectively. $^\ddagger$We employ an optimized prompt provided by the authors (see Subsection~\ref{prompt-sensitivity} in the Appendix for further details).}
\label{tab:all_results_test_large}
\vspace{-3mm}
\end{table*}

\subsubsection{Quality Assurance}
Before starting the quality assurance process, we first identify common issues that arise during the annotation and develop automated rules to detect easily identifiable annotation errors, such as incorrect string formatting. Annotators are then asked to correct these errors. To further ensure data quality and validity, we conduct several rounds of quality assurance. Initially, we focus on image quality by removing instances where images are blurry, dark, or contain distracting elements such as people or other dishes. We also verify image licenses by cross-referencing them with information on Wikimedia Commons. Next, we refine the dish categorization and descriptions, ensuring consistency in category assignments and maintaining descriptions free from ``information breaches'' (e.g., excluding regional details from the description). We standardize cuisine names and eliminate any redundancies. Finally, we meticulously review all country and area information to ensure its accuracy. This comprehensive approach guarantees the integrity and reliability of our dataset.

\subsubsection{Data Compilation}
In this phase, we verify the overall quality check done by annotators, and identify any potential inconsistencies that are missed during the quality assurance. Then, we compile the dataset by collecting the metadata into a single file. 

\subsection{VQA Generation}
In this phase, we generate VQA data by sampling from $\kbname$. An entry of VQA data comprises visual image, question text, and answer text. This process involves four stages: (1) conducting a similarity search for dish names, (2) constructing questions and contexts, (3) translating these elements into multiple languages, and (4) generating the VQA triplets.

\subsubsection{Dish Names Similarity Search}\label{sec:dish-name-similarity}
To identify similar dishes in our dataset, we follow the approach from~\citet{winata2024miners} to employ a multilingual model E5$_\text{LARGE}$ Instruct~\cite{wang2024multilingual} for computing text embedding.
Formally, given a dish $x$ with name $x_{\text{name}}$ and text description $x_{\text{desc}}$, we use a multilingual model $\theta$ to compute the embedding vector $v_x = \theta(\{x_\text{name};x_\text{desc}\})$, 
then apply cosine similarity to compute a score $s = \text{similarity}(v_i, v_j)$ between dish $i$ and dish $j$. 
For each dish, we consider the top-$k$ most similar dishes to generate distractors in the multiple choice question.

\subsubsection{Question and Context Construction}
\label{sec:question_context_construction}
Dish name prediction (Task 1) is divided into three question variations depending on the context:
(1a) \emph{no-context question}, where we simply ask for the name of the dish without any provided context;
(1b) \emph{contextualized question} where we provide additional information related to cuisine or location; and
(1c) \emph{adversarial contextualized question} which are similar to the contextualized questions but may include misleading location information to assess the model's robustness to irrelevant details.

For example, consider \textit{coxinha} from \textbf{Brazil}, shown in Figure~\ref{fig:tasks} (1b). A query with additional context here would be: \texttt{``What is the common name for this dish in \textbf{Brazil}?"} Here, the origin of \textit{coxinha}, \textbf{Brazil}, serves as the context. In contrast, adversarial context involves providing misleading or irrelevant information in terms of location or type of cuisine to assess the model’s robustness to such distractions. For instance, in the case of eggs benedict shown in Figure~\ref{fig:tasks} (1c), an adversarial context would be: \texttt{“Yesterday I had a nice lunch at a \textbf{Korean} restaurant. I am about to have this dish now. What is this dish called?”} In this scenario, the model should ignore the irrelevant detail \texttt{(“nice lunch at a Korean restaurant”)} and focus solely on the image and the question.

Only basic question without any provided context is available for regional cuisine prediction (Task 2). The data statistics for each task are presented in Table~\ref{tab:vqa_splits}.

\begin{figure*}[!th]
    \centering
    \begin{subfigure}[t]{\textwidth}
        \centering
        \includegraphics[width=0.97\linewidth]{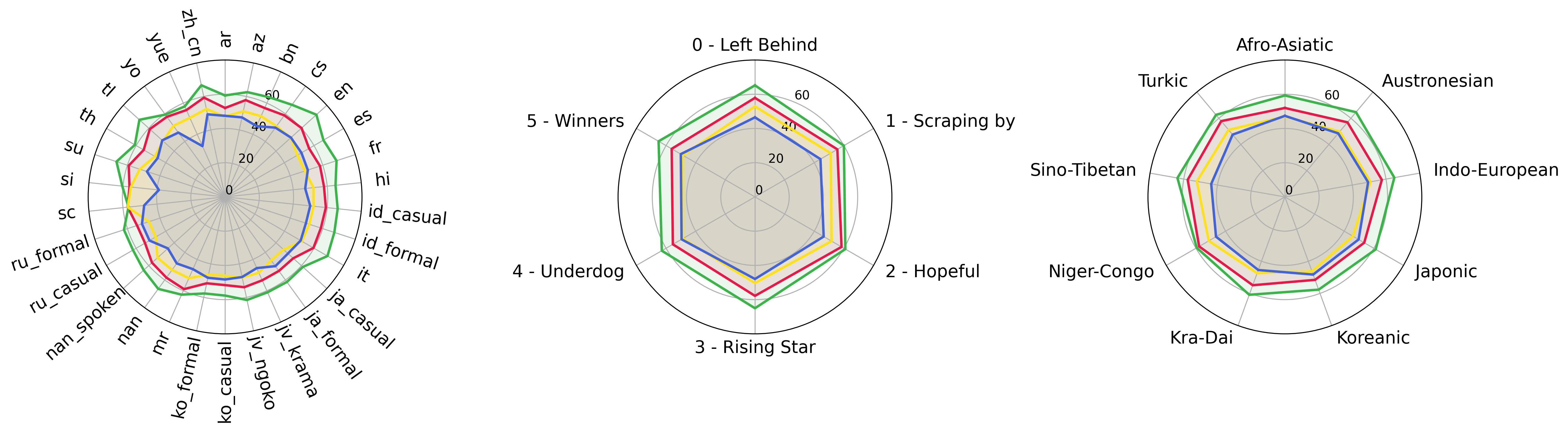}
        \caption{Multiple-choice question (MCQ).}
        \label{fig:all_radar_results_mcq}
    \end{subfigure}
    \begin{subfigure}[t]{\textwidth}
        \centering
        \includegraphics[width=0.97\linewidth]{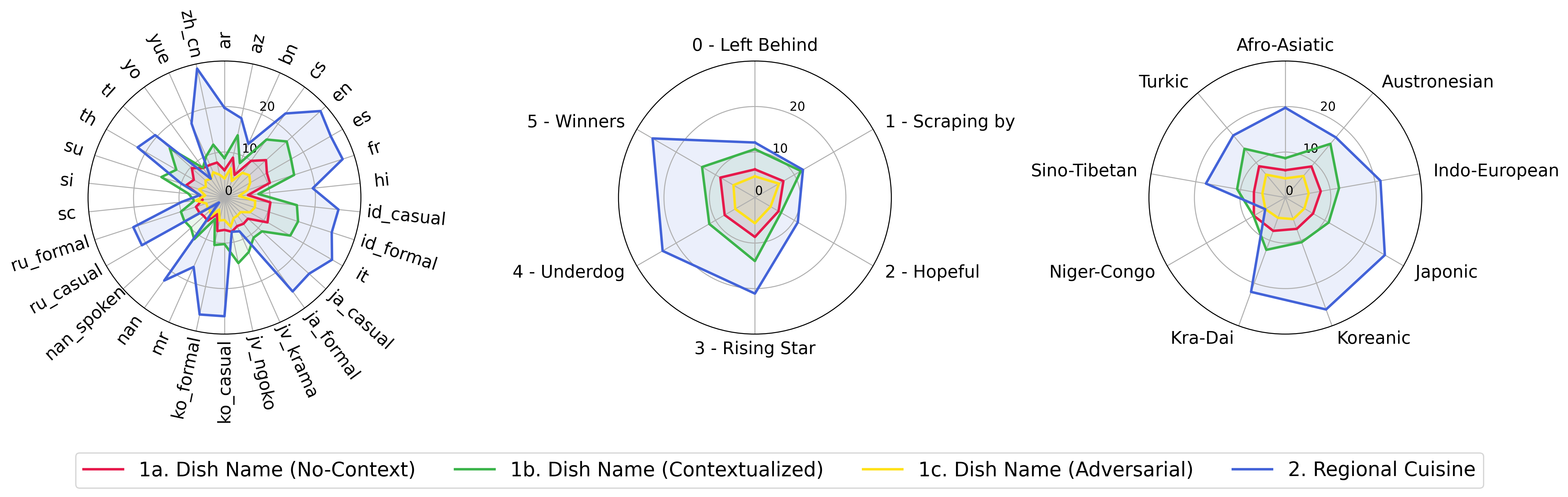} 
        \caption{Open-ended question (OEQ).}
        \label{fig:all_radar_results_oeq}
    \end{subfigure}
    \caption{Accuracy (\%) categorized by language \textbf{(left)}, language vitality \textbf{(center)}, and language family \textbf{(right)}. We classify the language vitality by following the classification proposed by~\citet{joshi2020state}.}
    \label{fig:all_radar_results}
\end{figure*}

\subsubsection{Multiple Language Translation}

\paragraph{Question and Context.}
All questions and contexts are initially collected in English, which are then carefully translated by native human speakers into 30 language varieties: 23 different languages with 7 languages having two different varieties each. We instructed the translators to prioritize the naturalness, and then followed by the diversity of translations when the duplication occurs.

\paragraph{Food Name Alias.}
Using Wikipedia pages as our primary source, we can verify if the English page has translations available in other languages. This enables us to extract dish names in multiple languages and compile them as translations for each dish. We utilize both the Wikipedia page titles in various languages and the alias text found on the English page. These translations are especially valuable for multilingual prompt translation, as they allow us to use the dish's native name instead of its English equivalent, enhancing cultural relevance and accuracy. We use the English name as default when the translation is unavailable.

\paragraph{Locations and Cuisines.}
As there are more than 400 unique locations, including countries, cities, and areas, we first translate the English locations into other languages by using GPT-4o, followed by proofreading each translation by the native speakers. The string values for the regional cuisines, i.e., the adjective form of the location in English, are translated in the same manner as location.

\paragraph{Morphological Inflections.}
Indo-European languages, such as Czech or Spanish, are rich in inflectional morphology which involves word modification to express different grammatical categories, such as number, gender, or case. For example, the equivalents of the phrases ``in Japan'' and ``from Japan'' in Czech are ``v~Ja\-pon\-sk\emph{u}'' and ``z~Ja\-pon\-sk\emph{a}'', respectively. We provide a framework for the human translators to use the inflections in the prompt template to prioritize the naturalness while keeping the inflections as few as possible.

\subsubsection{Generating VQA Triplets}
To ensure no overlap in train and test subsets, we split the dishes and the multilingual-questions into two subsets each, to ensure no dish or multilingual questions leakage between train and test. For every subset, we apply random sampling to get a pair of dish and its multilingual-questions. We use the dish entry in our WorldCuisines KB dataset to pick the image and the location to be injected to the context, if any. The answer candidates for multiple-choice were picked by utilizing similarity search (Section \ref{sec:dish-name-similarity}). We repeat this process until we reach the desired number of training or test samples, or until all possible dish and question combinations are used, discarding any duplicates.

\section{Experiments}

\subsection{Experimental Setup}

\paragraph{Metrics.}
We use accuracy as the primary metric to evaluate predictions. For Task 2 (open-ended), we employ BERTScore~\cite{zhang2019bertscore} with XLM-R Large~\cite{conneau2019cross} as a secondary metric to determine if the model-generated content includes food names similar to those in the gold labels. For open-ended questions, we compute the accuracy of each test sample against multiple references, including translations of the dish in different languages. This approach allows us to accommodate predictions that may not be in the expected language.

\paragraph{Models.}
We evaluate our benchmark on various available VLMs, including 15 open-source models and 3 proprietary models. During the inference of the open-source model, we use 16-bit floating point and employ greedy decoding. We access the proprietary models through API. The complete list of the models is available in Table \ref{tab:all_results_test_large}.

\begin{figure*}[!t]
    \centering
    \begin{subfigure}[t]{.49\linewidth}
        \centering\includegraphics[width=\linewidth, trim={{2.em} 0 0 0},clip]{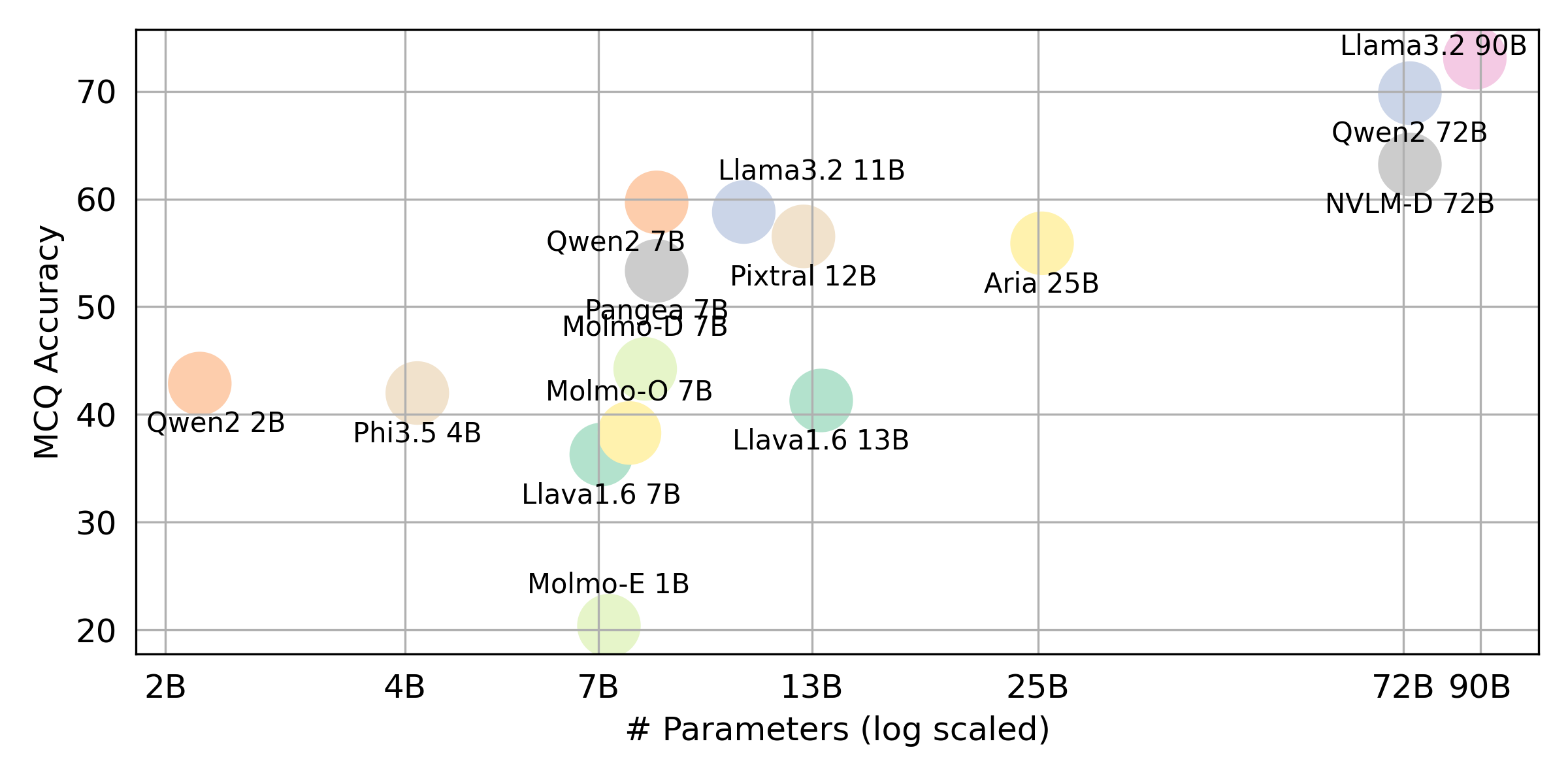}
        \caption{MCQ Accuracy vs. Parameters.}\label{fig:model_parameters_mcq}\smallskip
    \end{subfigure}
    \begin{subfigure}[t]{.49\linewidth}
        \centering\includegraphics[width=\linewidth, trim={{2.em} 0 0 0},clip]{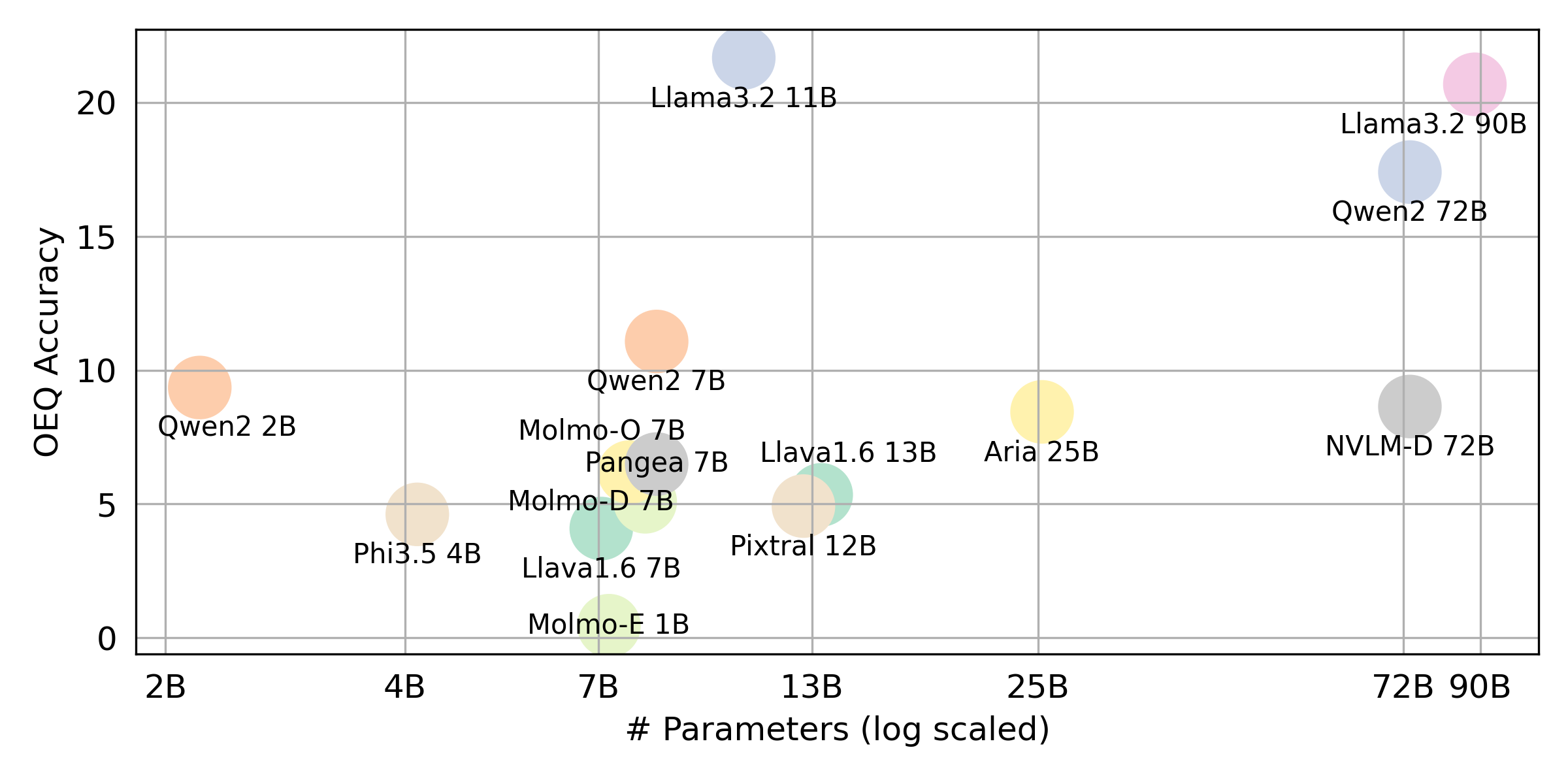}
        \caption{OEQ Accuracy vs. Parameters.}\label{fig:model_parameters_oeq}
    \end{subfigure}
    \caption{Scaling matters for MCQ~(\ref{fig:model_parameters_mcq}) and OEQ~(\ref{fig:model_parameters_oeq}).}
    \label{fig:model_parameters}
\end{figure*}

\section{Results and Discussion}

\subsection{Overall Results}
The results for $\vqaname$ are summarized in Table~\ref{tab:all_results_test_large}. The multiple-choice question (MCQ) results without any context exhibit significant variability, ranging from 30\% to 80\%, indicating considerable differences in model performance. This variability indicates that predicting MCQs remains a challenging task for many models. Notably, proprietary models, particularly GPT-4o, demonstrate exceptional performance, outperforming all other models. In the open-ended question (OEQ) setting, the task proves even more difficult than the MCQ, with models achieving a maximum accuracy of less than 20\% for dish name predictions and slightly higher for location predictions when no context is provided. However, incorporating context enhances performance across all settings, highlighting that context effectively guides the models in making better predictions. Interestingly, when the adversarial context is introduced, it misleads the models, leading to incorrect predictions and adding further complexity to the task. Among the models evaluated, Llama 3.2 Instruct significantly outperforms other open-source model families, while Qwen2 performs relatively better than Llava 1.6 and Molmo, despite having smaller model sizes.

\subsection{The Role of Context}
For dish name prediction (Task 1), incorporating more relevant context significantly enhances performance across all language families. However, when adversarial context is introduced, performance drops significantly. The adversarial context included in the prompt significantly affects the prediction. Instead of relying solely on the image input, the model often sways and makes predictions based on incorrect location or cuisine information, even when the context is unrelated to the query. This observation is particularly intriguing, as it signifies that such prompts can shift the model's attention and influence its generation process.

\subsection{Results by Language}
In Task 1 with OEQ setting (Figure~\ref{fig:all_radar_results_oeq}), some languages with non-Latin scripts, such as Arabic, Korean, Japanese, and Marathi, tend to perform poorly, with the exception of Chinese.
For Task 2 with OEQ setting, most models struggle with Sino-Tibetan languages (i.e., Chinese, Cantonese, and Hokkien) and Niger-Congo languages (i.e., Yoruba). In contrast, the models demonstrate relatively strong performance with Japonic, Koreanic, Kra-Dai (i.e., Thai), and Turkic (i.e., Azerbaijani) languages. We also observe that answering OEQs in underrepresented languages remains particularly challenging for the models, as shown by the relatively lower results for the ``left behind'', ``scraping by'', and ``hopeful'' languages. Interestingly, lower performance in the OEQ does not necessarily translate to the lower performance in the MCQ setting (Figure~\ref{fig:all_radar_results_mcq}) where the performance gap between language categories is less pronounced. The gap between OEQ and MCQ, especially for underrepresented languages, suggests that the bottleneck might lie in the factors beyond cultural understanding, such as text generation capabilities.

\subsection{Scaling Law}
It is evident that large models perform better than smaller ones, showing the scaling law still exists in this experiment, as shown in Figure~\ref{fig:model_parameters}. It is very interesting to see the same trend across different model families (e.g., Llava, Qwen, and even GPT-4o series). However, it is pretty clear for open-source models, Llama 3.2 Instruct has the lead for overall performance, which may be due the coverage of multilingual data used in its training, although it is still unclear since there is no evidence or supporting information that can back up the finding. Regardless, NVLM-D model does not perform as good as their base model Qwen2 VL Instruct in our benchmark. One reason could be the NVLM model is highly tuned for English, but not in languages other than English.

\section{Related Work}

\paragraph{Cultural VQA.} Several prior studies have focused on developing culturally relevant VQA benchmarks, including FM-IQA~\cite{gao2015you}, MCVQA~\cite{gupta2020unified}, xGQA~\cite{pfeiffer2022xgqa}, MaXM~\cite{changpinyo2023maxm}, MTVQA~\cite{tang2024mtvqa}, MABL~\cite{kabra2023multi}, MAPS~\cite{liu2024multilingual}, and MaRVL~\cite{liu2021visually}. Additionally, CVQA~\cite{romero2024cvqa} and CulturalVQA~\cite{nayak2024benchmarking} provide VQA datasets that cover various regions and diverse topics, including food, with CVQA also offering questions in multiple languages alongside English translations. SEA-VQA~\cite{urailertprasert2024sea} specifically benchmarks the South East Asian region. In contrast, FoodieQA~\cite{li2024foodieqa} and World Wide Dishes~\cite{magomere2024you} are benchmark focusing on food. Our work is similarly motivated by using food as a cultural proxy, but it distinguishes itself with a significantly larger dataset and broader coverage of languages and cultures.

\paragraph{Multi-modal LLMs.} Recent advancements in VLMs have led to the emergence of multi-modal LLMs that can process both images and text. LLaVA~\cite{liu2024visual} exemplifies this approach by utilizing Vicuna~\cite{zheng2023judging} as an image encoder, thereby enhancing visual understanding. This architecture has set a precedent for other VLMs, including Qwen2-VL~\cite{bai2023qwen}, Llama 3.2~\cite{dubey2024llama}, Pixtral~\cite{agrawal2024pixtral}, Phi-3.5 Vision~\cite{abdin2024phi}, Molmo~\cite{deitke2024molmo}, Aria~\cite{li2024aria}, Pangea~\cite{yue2024pangea}, and NVLM~\cite{dai2024nvlm}, each leveraging their respective large language models for multi-modal tasks. In a specialized application, FoodLMM~\cite{yin2023foodlmm} focuses specifically on the food domain, training on publicly available food datasets and conversational data generated by GPT-4~\cite{achiam2023gpt}. Our work evaluates the capabilities of these models within the food domain, offering insights into their performance and potential applications in culinary-related tasks across multicultural settings.

\section{Conclusion}
We introduce $\benchmarkname$, an open-source, large-scale benchmark designed for multilingual and multicultural, visually grounded language understanding. It comprises over 1 million data points across 30 languages and dialects. Our findings reveal that this benchmark remains challenging for VLMs, particularly with dishes from specific regions and in low-resource languages. This provides insight into how well models understand regional cuisines. To enhance usability, we offer a dedicated evaluation split with two datasets of varying sizes. Our evaluation shows that while VLMs perform better with the correct context, they struggle with adversarial contexts intended to mislead them. Additionally, we are releasing a comprehensive knowledge base, VQA dataset, code, and leaderboard as open-source resources to support future research.

\section*{Acknowledgements}
We extend our gratitude to everyone who has supported our project, especially the numerous annotators who provided meticulous and comprehensive annotations and conducted thorough quality checks. Special thanks to Francesca Porcu for her assistance with the Sardinian language and to Shintaro Ozaki for his help with Japanese. We are also deeply appreciative of Nayeon Lee and Wenliang Dai for their insightful discussions and for integrating NVLM into our benchmark. Additionally, we thank Xiang Yue and Yueqi Song for their help in integrating Pangea into our benchmark.

\section*{Limitations}
In this paper, we limit our investigation to avoid exhaustively evaluating all possible models due to resource constraints. Our primary focus is on developing a benchmark that facilitates exploration for future research. We also provide a training data split for reference, allowing other researchers to utilize it to enhance their VLMs and evaluate their models against our test sets. Currently, we include 30 different languages and dialects, establishing one of the largest and most diverse benchmarks for comprehensive multilingual VQA. We aim to extend this benchmark to encompass additional languages in the future, making it more inclusive and representative of a broader range of linguistic diversity. 

It is important to note that our food entries are currently sourced from English Wikipedia. Although we aim to include as many diverse dishes as possible, we acknowledge that this approach limits the coverage of some regions. This is due to language affects commonsense and its specific knowledge \citep{sakai2024mcsqa}, which in turns suggesting insufficiency of sourcing only English Wikipedia. Nevertheless, our dataset serves as a valuable starting point. In future work, we plan to incorporate entries from non-English Wikipedia pages to improve regional representation and cultural diversity. For evaluation purposes, we include accuracy metrics for overall model performance and BERTScore for more detailed analysis. However, we recognize that evaluating VQA model performance on multicultural data remains an open challenge. Appropriate evaluation metrics are needed to effectively model the diversity of cultural contexts and linguistic variations. Addressing this issue will be a key focus of our future research efforts.

\section*{Ethical Considerations}
Our research focuses on evaluating VLMs within the context of multilingual and multicultural VQA, a field that holds significant implications for diverse multilingual communities. We are committed to conducting our data collection and evaluations with the highest standards of transparency and fairness. To achieve this, we have adopted a crowd-sourcing approach for the annotation process, inviting volunteers to contribute and become co-authors if they provide significant contributions. We follow the guidelines from ACL for authorship eligibility as shown in ~\url{https://www.aclweb.org/adminwiki/index.php/Authorship_Changes_Policy_for_ACL_Conference_Papers}. In line with our commitment to openness and collaboration, we will release our dataset under an open-source license, CC-BY-SA 4.0.

\bibliography{custom}
\bibliographystyle{acl_natbib}

\appendix

\begin{CJK}{UTF8}{ipxm}
\begin{table*}[!ht]
\centering
\resizebox{0.95\textwidth}{!}{
    \begin{tabular}{lcll}
    \toprule
    \textbf{Attribute} & \textbf{Value} & \textbf{Description} & \textbf{Example}\\ \midrule
    Name & String & Name of the dish. & ``Dorayaki'' \\
    Alias & List<Dict> & Name alias, i.e. the name in the original language. & [\{``どら焼き'': ``Japanese''\}] \\
    Coarse-grained categories & List<String> & Coarse-level categories. & [``Pancake'', ``Dessert''] \\
    Fine-grained categories & List<String> & Fine-level categories. & [``Wagashi Pancake''] \\
    Cuisines & String & Name of cuisine. & ``Japanese'' \\
    Associated Cuisines & String & Associated cuisines to the dish. & ``Japanese'' \\	
    Area & String & Specific region where the dish is originated & ``Ueno'' \\
    Countries & String & Specific region where the dish is originated & ``Japan''\\
    Region[1..5] & String & Specific continent where the dish is originated & ``Eastern Asia''\\
    \makecell[l]{Text Description} & \makecell[c]{String} & \makecell[l]{Short description of the dish, including the ingredients \\ used to prepare the dish or the cooking method.} & \makecell[[l]{``The dish consists of two small pancake-like \\ patties made from castella wrapped around \\ a filling of sweet bean paste.''}
    \\
    Image[1..8] URL & String & Image link to Wikimedia Commons. & ``\ldots/commons/9/9c/Dorayaki\_001\_(3).jpg'' (\includegraphics[width=0.06\textwidth]{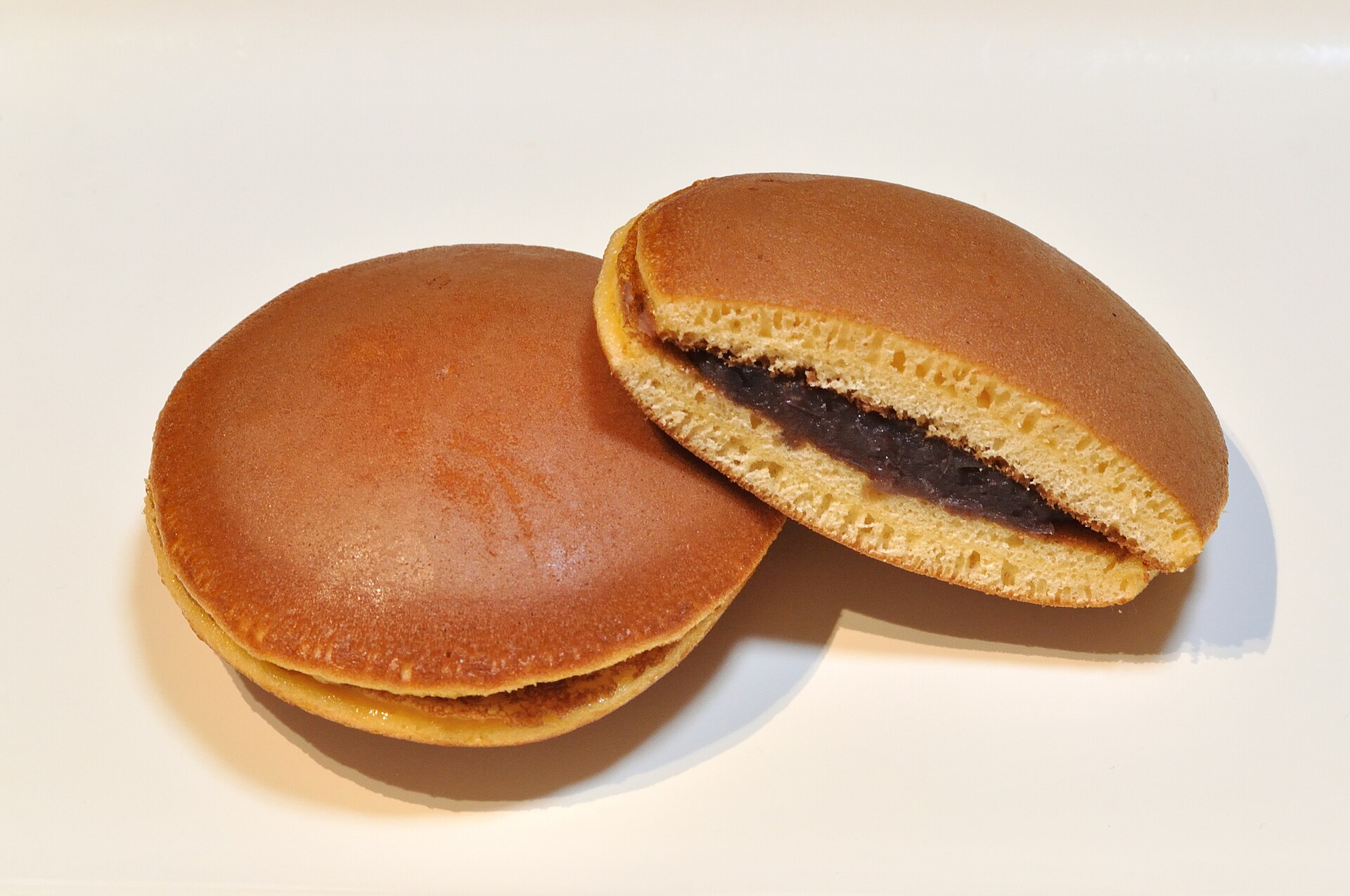}) \\
    Image[1..8] License & String & License of the image & ``CC BY-SA 3.0'' \\
    \bottomrule
    \end{tabular}
}
\caption{$\kbname$ attributes in $\benchmarkname$.}
\label{tab:attributes}
\end{table*}
\end{CJK}

\section{Data Statement}
\label{sec:data_statement}

\subsection{Executive Summary}
$\benchmarkname$ is a vision-language benchmark comprised of two resources: (1) $\vqaname$, a multilingual parallel question answering dataset covering 30 languages and dialects where each dish image is accompanied by questions and context constructed through human translation; and (2) $\kbname$, a knowledge base containing images and metadata associated with the dishes. 

\subsection{Curation Rationale}
The goal of $\benchmarkname$ is to evaluate the cultural understanding of vision-language models (VLMs) within the food domain. To achieve this, we develop $\vqaname$ and $\kbname$. Dish names and their information are collected from English Wikipedia, and the images are selected from Wikimedia Commons to ensure a permissive license, with an emphasis on representing a wide range of food categories and geographic origins (or where the dish is popular). This selection strategy aims to provide insights into the VLMs' ability to generalize across diverse culinary and cultural contexts.

\subsection{Language Variety}
$\benchmarkname$ covers 30 languages and dialects spoken across diverse countries and regions. The complete list of languages and dialects is shown in Table~\ref{tab:language_distribution}. An example of the multilingual prompt is shown in Table~\ref{tab:multilingual-prompt-example}.

\begin{table*}[!th]
    \centering
    \resizebox{.9\textwidth}{!}{
    \begin{tabular}{lcccc}
        \toprule
        \textbf{Language Name} & \textbf{Language Vitality}$^\dagger$ & \textbf{Resource Classification}$^\ddagger$ & \textbf{Linguistic Register} & \textbf{Additional Notes} \\
        \midrule
        \textbf{Austronesian}\\
            \multirow{2}{*}{$\quad$Indonesian} & \multirow{2}{*}{Institutional} & \multirow{2}{*}{3 - Rising Star} & Formal &  \\
             & & & Casual & \\
            $\quad$Tagalog & Institutional & 3 - Rising Star & & \\
            $\quad$Sundanese & Stable & 1 - Scraping by & Loma & Common speech form \\
            \multirow{2}{*}{$\quad$Javanese} & \multirow{2}{*}{Institutional} & \multirow{2}{*}{1 - Scraping by} & Krama & Central-Java dialect, polite form \\
             & & & Ngoko & Central-Java dialect, casual form \\ \midrule
        \textbf{Japonic}\\
            \multirow{2}{*}{$\quad$Japanese} & \multirow{2}{*}{Institutional} & \multirow{2}{*}{5 - Winners} & Formal & Polite form or teinei-go \\
             & & & Casual & Daily conversation \\ \midrule
        \textbf{Sino-Tibetan}\\
            $\quad$Chinese & Institutional & 5 - Winners & & Standard Mandarin \\
            $\quad$Cantonese & Institutional & 1 - Scraping by & & \\
            \multirow{2}{*}{$\quad$Hokkien} & \multirow{2}{*}{Institutional} & \multirow{2}{*}{0 - Left Behind} & Written & Medan dialect \\
              & & & Spoken & Medan dialect \\ \midrule
        \textbf{Koreanic}\\
            \multirow{2}{*}{$\quad$Korean} & \multirow{2}{*}{Institutional} & \multirow{2}{*}{4 - Underdog} & Formal & \\
             & & & Casual & \\ \midrule
        \textbf{Kra-Dai}\\
            $\quad$Thai & Institutional & 3 - Rising Star & & \\ \midrule
        \textbf{Indo-European}\\
            $\quad$English & Institutional & 5 - Winners & & \\
            $\quad$Spanish & Institutional & 5 - Winners & & Latin-American dialect \\
            $\quad$French & Institutional & 5 - Winners & & \\
            \multirow{2}{*}{$\quad$Russian} & \multirow{2}{*}{Institutional} & \multirow{2}{*}{4 - Underdog} & Formal & \\
             & & & Casual & \\
            $\quad$Czech & Institutional & 4 - Underdog & & \\
            $\quad$Italian & Institutional & 4 - Underdog & & \\
            $\quad$Hindi & Institutional & 4 - Underdog & & \\
            $\quad$Bengali & Institutional & 3 - Rising Star & & \\
            $\quad$Marathi & Institutional & 2 - Hopeful & & \\
            $\quad$Sardinian & \emph{Endangered} & 1 - Scraping by & & Logudorese (src) \\
            $\quad$Sinhala & Institutional & 0 - Left Behind & Formal & Spoken form \\ \midrule
        \textbf{Afro-Asiatic}\\
            $\quad$Arabic (MSA) & Institutional & 5 - Winners & & \\ \midrule
        \textbf{Niger-Congo}\\ 
            $\quad$Yoruba & Institutional & 2 - Hopeful & & \\ \midrule
        \textbf{Turkic}\\
            $\quad$Azerbaijani & Institutional & 1 - Scraping by & & North Variety (azj) \\
        \bottomrule
    \end{tabular}
    }
    \caption{The details of languages used in the prompt generation for our VQA dataset. $^\dagger$Taken from Ethnologue \citep{campbell2008ethnologue}. $^\ddagger$Based on \citet{joshi2020state}.}
    \label{tab:language_distribution}
\end{table*}

\begin{CJK}{UTF8}{ipxm}
\begin{table*}
    \centering
    \resizebox{0.95\textwidth}{!}{
    \begin{tabular}{lcc|cc}
        \toprule
        \multirow{2}{*}{\textbf{Language}} & \multicolumn{2}{c|}{\textbf{Question Prompt}} & \multicolumn{2}{c}{\textbf{Answer}} \\
        & Multi-choice question (MCQ) & Open-ended question (OEQ) & ID & Text \\
        
        \midrule
        \makecell[l]{English}
        & \makecell[l]{Yesterday I had a nice lunch at a Japanese restaurant.\\ I am about to have this dish now. What is this dish called?\\ 1. Hangtown fry\\ 2. Zucchini slice\\ 3. Chawanmushi\\ 4. Rolex\\ 5. Egg foo young\\ \\ Print only the answer with a single answer id (1,2,3,4,5).}
        & \makecell[l]{Yesterday I had a nice lunch at a Japanese restaurant.\\ I am about to have this dish now. What is this dish called?\\ \\ Print only the answer.}
        & 5 & Egg foo young \\


        \midrule
        \makecell[l]{French}
        & \makecell[l]{Hier, j'ai pris un bon déjeuner dans un restaurant japonais.\\ Je suis sur le point de manger ce plat maintenant.\\ Comment appelle-t-on ce plat ?\\ 1. Hangtown fry\\ 2. Zucchini slice\\ 3. Chawanmushi\\ 4. Rolex\\ 5. Fu yung hai\\ \\ Print only the answer with a single answer id (1,2,3,4,5).}
        & \makecell[l]{Hier, j'ai pris un bon déjeuner dans un restaurant japonais.\\ Je suis sur le point de manger ce plat maintenant.\\ Comment appelle-t-on ce plat ?\\ \\ Print only the answer.}
        & 5 & Fu yung hai \\
        
        \midrule
        \makecell[l]{Indonesian\\(Formal)}
        & \makecell[l]{Kemarin, saya menyantap makan siang yg nikmat di restoran Jepang.\\ Sekarang saya akan menyantap hidangan ini.\\ Disebut apakah hidangan ini?\\ 1. Hangtown fry\\ 2. Zucchini slice\\ 3. Chawanmushi\\ 4. Rolex\\ 5. Puyunghai\\ \\ Print only the answer with a single answer id (1,2,3,4,5).}
        & \makecell[l]{Kemarin, saya menyantap makan siang yg nikmat di restoran Jepang.\\ Sekarang saya akan menyantap hidangan ini.\\ Disebut apakah hidangan ini?\\ \\ Print only the answer.}
        & 5 & Puyunghai \\
        
        \midrule
        \makecell[l]{Indonesian\\(Casual)}
        & \makecell[l]{Kemarin aku makan siang enak di restoran Jepang.\\ Sekarang mau makan makanan ini.\\ Makanan ini disebut apa?\\ 1. Hangtown fry\\ 2. Zucchini slice\\ 3. Chawanmushi\\ 4. Rolex\\ 5. Puyunghai\\ \\ Print only the answer with a single answer id (1,2,3,4,5).}
        & \makecell[l]{Kemarin aku makan siang enak di restoran Jepang.\\ Sekarang mau makan makanan ini.\\ Makanan ini disebut apa?\\ \\ Print only the answer.}
        & 5 & Puyunghai \\
        
        \midrule
        \makecell[l]{Japanese\\(Formal)}
        & \makecell[l]{昨日、私は日本料理店で美味しい昼食を食べました。\\今まさにこの料理を食べようとしています。\\この料理の名前は何ですか?\\ 1. Hangtown fry\\ 2. Zucchini slice\\ 3. 茶碗蒸し\\ 4. Rolex\\ 5. 芙蓉蛋\\ \\ Print only the answer with a single answer id (1,2,3,4,5).}
        & \makecell[l]{昨日、私は日本料理店で美味しい昼食を食べました。\\今まさにこの料理を食べようとしています。\\この料理の名前は何ですか?\\ \\ Print only the answer.}
        & 5 & 芙蓉蛋 \\

        \midrule
        \makecell[l]{Japanese\\(Casual)}
        & \makecell[l]{昨日日本料理のお店で美味しいランチを食べたんだけど、\\今まさに食べてるこの料理の名前は何？\\ 1. Hangtown fry\\ 2. Zucchini slice\\ 3. 茶碗蒸し\\ 4. Rolex\\ 5. 芙蓉蛋\\ \\ Print only the answer with a single answer id (1,2,3,4,5).}
        & \makecell[l]{昨日日本料理のお店で美味しいランチを食べたんだけど、\\今まさに食べてるこの料理の名前は何？\\ \\ Print only the answer.}
        & 5 & 芙蓉蛋 \\

        

        \midrule
        \makecell[l]{Javanese\\(Krama)}
        & \makecell[l]{Kaping wingi kula nedha nikmat ing restoran Jepang.\\ Kula kepengin nedha menika malih sakmenika.\\ Naminipun nopo dhaharan menika?\\ 1. Hangtown fry\\ 2. Zucchini slice\\ 3. Chawanmushi\\ 4. Rolex\\ 5. Endhog foo young\\ \\ Print only the answer with a single answer id (1,2,3,4,5).}
        & \makecell[l]{Kaping wingi kula nedha nikmat ing restoran Jepang.\\ Kula kepengin nedha menika malih sakmenika.\\ Naminipun nopo dhaharan menika?\\ \\ Print only the answer.}
        & 5 & Endhog foo young \\
        
        \midrule
        \makecell[l]{Javanese\\(Ngoko)}
        & \makecell[l]{Wingi aku mangan enak ndek restoran Jepang.\\ Aku pengen mangan neh saiki.\\ Opo jenenge panganan iki?\\ 1. Hangtown fry\\ 2. Zucchini slice\\ 3. Chawanmushi\\ 4. Rolex\\ 5. Endhog foo young\\ \\ Print only the answer with a single answer id (1,2,3,4,5).}
        & \makecell[l]{Wingi aku mangan enak ndek restoran Jepang.\\ Aku pengen mangan neh saiki.\\ Opo jenenge panganan iki?\\ \\ Print only the answer.}
        & 5 & Endhog foo young \\


        

        \bottomrule
    \end{tabular}
    }
    \caption{Multilingual prompt example of Task 1 (c) adversarial in 8 language variants (out of 30). The visual image given is an image of Egg foo young, a Chinese cuisine. The ``\texttt{qa\_id}'' of this example is 1806.}
    \label{tab:multilingual-prompt-example}
\end{table*}
\end{CJK}

\begin{figure*}[!th]
    \centering
    \includegraphics[width=\linewidth]{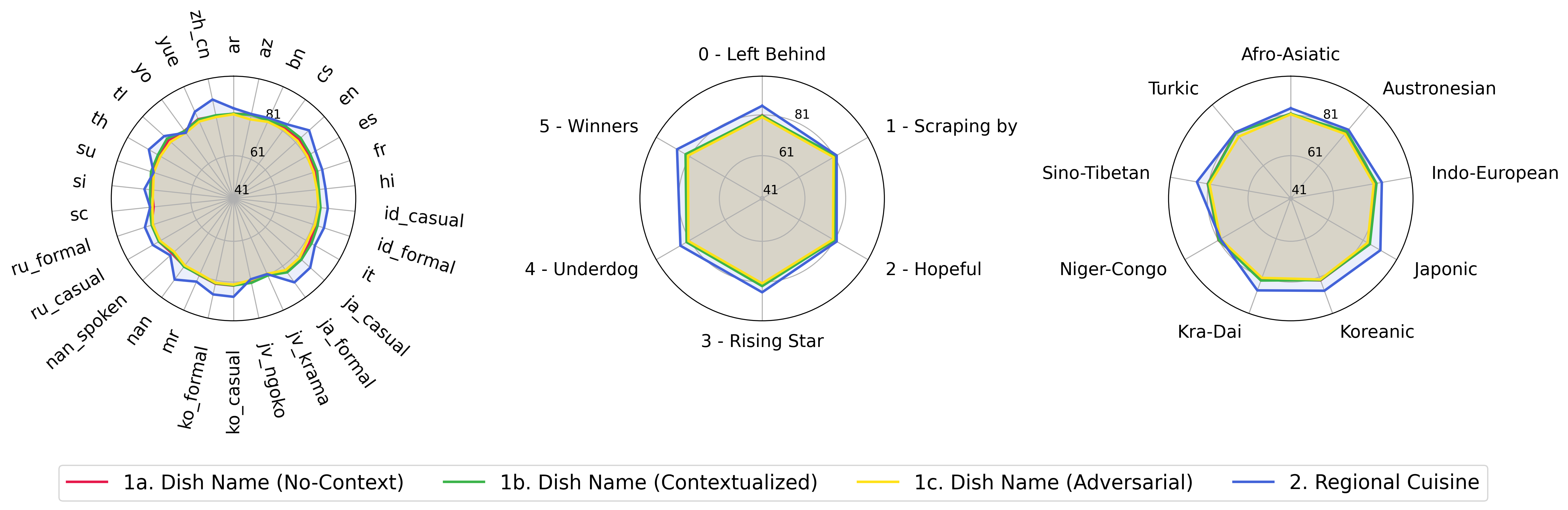}
    \caption{BERTScore (\%) categorized by language \textbf{(left)}, language vitality \textbf{(center)}, and language family \textbf{(right)}. We classify the language vitality by following the classification from~\citet{joshi2020state}.}
    \label{fig:bert_score_radar}
\end{figure*}

\begin{figure}[!t]
    \centering
    \includegraphics[width=\linewidth]{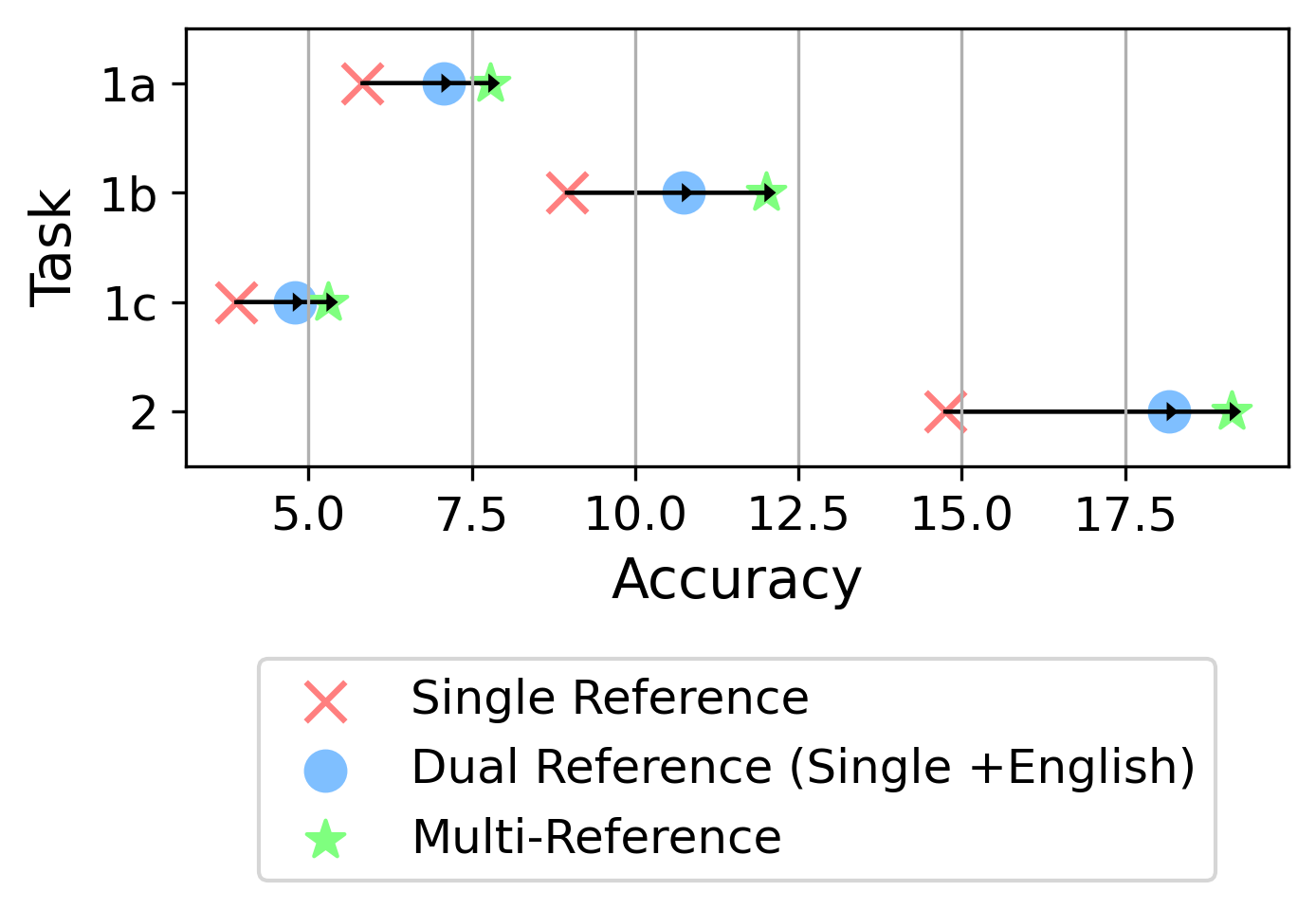}
    \caption{Model performance evaluated with different references on open-ended question.}
    \label{fig:oe_reference_results}
\end{figure}

\subsection{Annotator Demographic}
Over 30 annotators are involved in building $\benchmarkname$, specifically in translating the query and context for the $\vqaname$ dataset. Most annotators are native speakers of the target languages or dialects included in our data; some are L2 speakers with more than 10 years of study in their respective languages. The detailed demographics for each language are elaborated below.
\subsubsection{Austronesian}
\paragraph{Indonesian}
Two native Indonesian speakers are involved as translators. One is in the 26--35 age range, and the other is in the 16--25 age range.

\paragraph{Tagalog}
One native Tagalog speaker in the 16--25 age range is involved as a translator.

\paragraph{Sundanese}
Two L2 Sundanese speakers contribute to the translation. One, in the 16--25 age range with 15 years of experience with the Sundanese language, assists with translation. The other, in the 26--35 age range with 25 years of experience with the language, primarily serves as the proofreader.

\paragraph{Javanese}
One native Javanese speaker with Central Java dialect in the 16--25 age range translates for both registers of the language (Krama and Ngoko).

\subsubsection{Japonic}
\paragraph{Japanese}
Three L2 Japanese speakers with over 10 years of language study contribute to the Japanese translation. Two are in the 26--35 age range, and one is in the 36--45 age range. A native Japanese speaker then proofreads the translated sentences. Additionally, one native Japanese speaker from Western Japan in the 16--25 age range gives input for the casual form.

\subsubsection{Sino-Tibetan}
\paragraph{Chinese}
One native Chinese speaker in the 16--25 age range is involved as a translator.

\paragraph{Cantonese}
Two native Cantonese speakers are involved as translators. One is in 36--45 age range, and the other is in the 16--25 age range.

\paragraph{Hokkien}
Two native Hokkien speakers in the Medan dialect translate for both written and spoken versions of the language. Both are in the 26--35 age range.

\subsubsection{Koreanic}
\paragraph{Korean}
One native Korean speaker in the 16--25 age range translates the formal and casual versions of the language.

\subsubsection{Kra-Dai}
\paragraph{Thai}
One native Thai speaker in the 26--35 age range is involved as a translator.

\subsubsection{Indo-European}
\paragraph{English}
Query and context in English are constructed. All are L2 English speakers with over 20 years of study and have lived in the English speaking countries. Four of the annotators are in the 26--35 age range, and one is in 36--45 age range. Two native English speakers skimmed through the prompt templates.

\paragraph{Spanish}
One native Spanish speaker in the 26--35 age range translates the Latin-American versions or dialects of the language.

\paragraph{French}
One native French speaker and one L2 speaker are involved as translators. The native speaker is in the 26--35 age range, and the L2 speaker is in the 36--45 age range.

\paragraph{Russian}
One native Russian speaker in the 26--35 age range is involved as translators. One L2 speaker in 36--45 proofreads the template for inflection.

\paragraph{Czech}
One native Czech speaker in the 36--45 age range is involved as a translator.

\paragraph{Italian}
Two native Italian speakers, both in the 36--45 age range, are involved as translators.

\paragraph{Hindi}
One native Hindi speaker in the 26--35 age range is involved as a translator.

\paragraph{Bengali}
One native Bengali speaker in the 26--35 age range is involved as a translator.

\paragraph{Marathi}
One native Marathi speaker in the 26--35 age range is involved as a translator.

\paragraph{Sardinian}
One native Logudorese Sardinian speaker in the 36--45 age range is involved as a translator.

\paragraph{Sinhala}
One native Sinhala speaker in the 26--35 age range is involved as a translator.

\subsubsection{Afro-Asiatic}
\paragraph{Arabic (MSA)}
One native Arabic speaker in the 26--35 age range is involved in the Modern Standard Arabic (MSA) translation.

\subsubsection{Niger-Congo}
\paragraph{Yoruba}
One native Yoruba speaker in the 16--25 age range is involved as a translator.

\subsubsection{Turkic}
\paragraph{Azerbaijani}
One native Azerbaijani speaker in the 16--25 age range is involved as a translator.

\begin{figure*}[!t]
    \centering
    \includegraphics[width=0.7\linewidth]{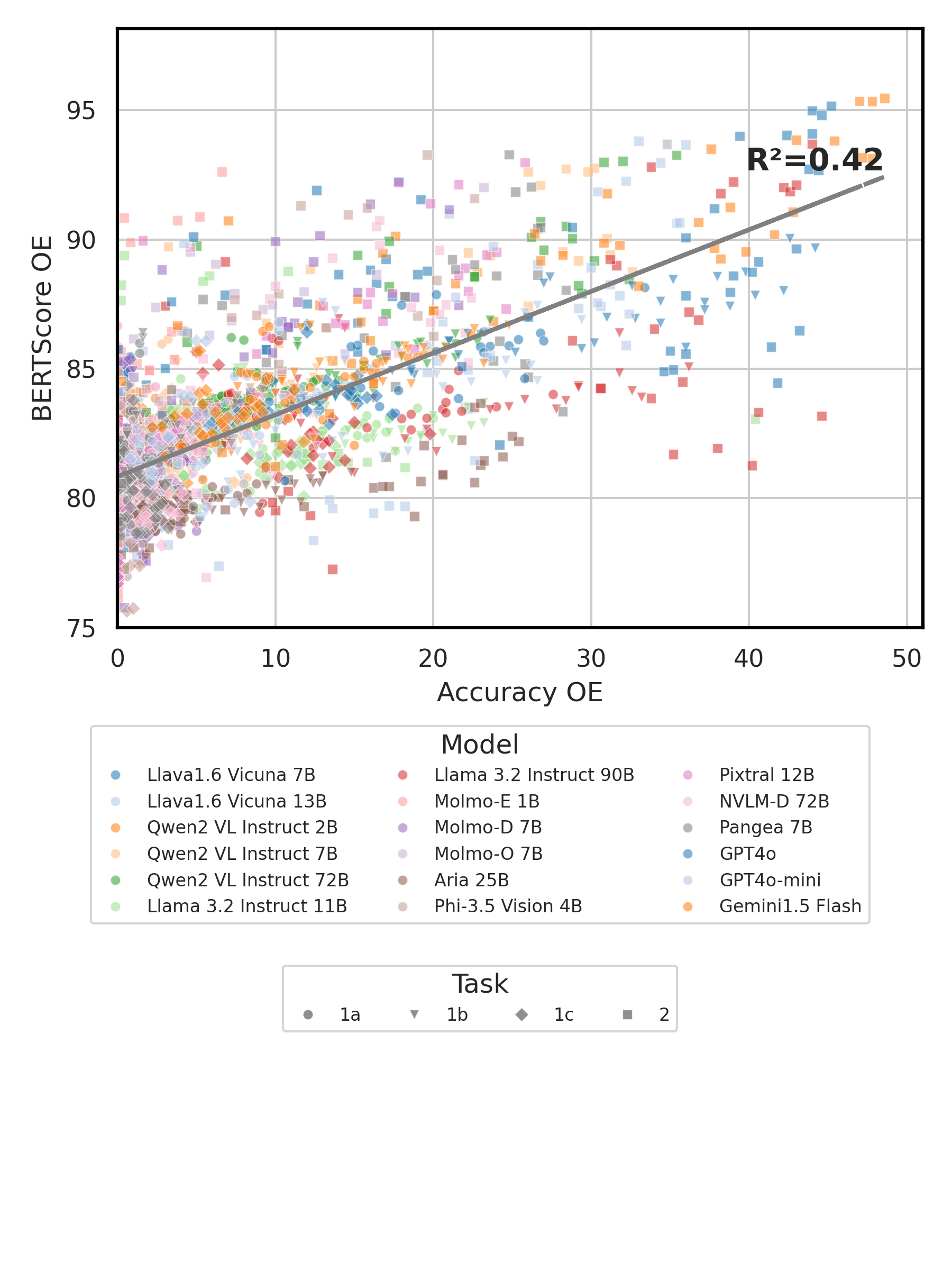}
    \caption{Regression Analysis for BERTScore OE vs. Accuracy OE.}
    \label{fig:bert_score_scatter}
\end{figure*}

\section{Open-Source Collaborative Effort}
The $\benchmarkname$ data collection and benchmark construction is a fully open-source project. We invite contributions from researchers, practitioners, and grassroots communities, such as local NLP communities, who are interested in participating. Contributions can include data collection, annotation, quality checks, and evaluation. To ensure high-quality data, we engage native speakers of local languages in the annotation process with strict quality control (QC). The contributors who provide substantial contribution are invited to have co-authorship on this paper. We follow the guidelines from ACL for authorship eligibility.\footnote{The ACL guidelines can be found at~\url{https://www.aclweb.org/adminwiki/index.php/Authorship_Changes_Policy_for_ACL_Conference_Papers}.} Our goal is to develop a resource and benchmark that will have a meaningful impact on future research. To achieve this, we are dedicated to expanding language coverage and ensuring that contributions are as \emph{inclusive and diverse} as possible.

\section{Detailed Dataset Construction}
\subsection{Dataset Compilation}
Our dataset, comprising 2,414 dishes and 6,084 images, was meticulously compiled and verified manually. Key metadata includes dish name, alias, coarse- and fine-grained categories, cuisines, regions, descriptions, images, and their licenses. The compilation process followed these steps:
\begin{itemize}
\item We listed dish names for annotation.
\item Annotators filled metadata fields and selected up to 8 licensed images per dish, guided by instruction documentation with examples for consistent accuracy.
\item Post-annotation, annotators formed subgroups to verify specific metadata categories, ensuring detailed and consistent data across fields such as categories, cuisines, regions, descriptions, and images. 
\end{itemize}

\subsection{Negative Sampling}

Recall that from our annotations, we have detailed metadata for all 2,414 dishes, including the dish name, coarse-grained categories, fine-grained categories, countries, and text descriptions. The negative answers were sampled using the following procedure:

\begin{itemize}
    \item[(1)] We used a multilingual model, specifically E5-LARGE Instruct, to compute the text embeddings. Each embedding was generated by concatenating the dish name with its corresponding text description.
    \item[(2)] To identify negative samples, we computed the cosine similarity between the embeddings of the target dish and those of all other dishes in the dataset. The top-K most similar dishes were selected under three different conditions:
        \begin{itemize}
            \item Same Fine-grained Category: Select top-K dishes from the same fine-grained category as the target dish.
            \item Same Coarse-grained Category: Select top-K dishes from the same coarse-grained category but potentially different fine-grained categories.
            \item No Category Restriction: Select top-K dishes from the entire dataset without any restriction on categories.
        \end{itemize}       
    Here, we used K=15, resulting in 45 candidate dishes in total.
    \item[(3)] Each MCQ consists of five options: one correct answer and four negative answers. The negative answers were chosen as follows:
    \begin{itemize}
        \item Two Difficult Options: The first two negative answers were selected from dishes in the same fine-grained category. These are intended to be more challenging for the model to distinguish.
        \item One Medium Option: The third negative answer was selected from dishes in the same coarse-grained category.
        \item One Easy Option: The fourth negative answer was selected from dishes without any category restriction, making it likely to be easier to identify as incorrect.
    \end{itemize}
    This approach ensures a balanced difficulty among the negative options, with two difficult, one medium, and one easy negative answer.
    \item[(4)] \textit{(Optional)} Specifically for task 2, where the question involves identifying the correct location (country) of a dish, we followed a slightly modified approach:
    \begin{itemize}
        \item From the previously retrieved 4 negative options, we identified the countries associated with each dish.
        \item We then excluded the countries that are valid locations for the correct dish. The remaining countries were used to create the negative options for the location-based question.
    \end{itemize}
\end{itemize}

\section{More Results}

\subsection{Primary Metric: Accuracy (\%)}
Table~\ref{tab:all_results_accuracy} presents the comprehensive results of $\vqaname$ for both Test Small and Test Large. Additionally, we examine the performance gap between different references used in the evaluation, with the results displayed in Figure~\ref{fig:oe_reference_results}.

\subsection{Secondary Metric: BERTScore}
As a secondary metric, we employ BERTScore using XLM-R Large as the base model. Table~\ref{tab:all_results_bertscore} presents the comprehensive results of $\vqaname$ for both Test Small and Test Large. Figure~\ref{fig:bert_score_radar} illustrates the model's performance categorized by language, language vitality, and language family.

\begin{table*}[!th]
\centering
\resizebox{0.91\textwidth}{!}{
    \begin{tabular}{lrrrrrr|rr|rr}
    \toprule
    \multicolumn{1}{l}{\multirow{3}{*}{\textbf{Model (Accuracy \%)}}} & \multicolumn{6}{c|}{\textbf{Task 1 (Dish Name)}} & \multicolumn{2}{c|}{\multirow{2}{*}{\textbf{\makecell{Task 2 \\ (Location)}}}} & \multicolumn{2}{c}{\multirow{2}{*}{\textbf{Average}}} \\ 
    & \multicolumn{2}{c}{\textbf{(a) no-context}} & \multicolumn{2}{c}{\textbf{(b) contextualized}} & \multicolumn{2}{c|}{\textbf{(c) adversarial}} & & \\
    & \hspace{.8em}MCQ & \multicolumn{1}{c}{OEQ}  & \hspace{2em}MCQ & \multicolumn{1}{c}{OEQ} & \hspace{.8em}MCQ & \multicolumn{1}{c|}{OEQ} & \hspace{.8em}MCQ & \multicolumn{1}{c|}{OEQ} & \hspace{.5em}MCQ & \multicolumn{1}{c}{OEQ} \\ \midrule
    \textbf{Test Small (12k)} \\ \midrule
    \textbf{Open-Source} & & & & & & & & & & \\
    $\quad$Llava1.6 Vicuna 7B & 33.63 & 0.87 & 43.13 & 2.83 & 28.67 & 0.60 & 27.77 & 7.93 & 33.30 & 3.06 \\
    $\quad$Llava1.6 Vicuna 13B & 40.87 & 1.00 & 50.30 & 4.17 & 38.37 & 1.60 & 31.07 & 8.63 & 40.15 & 3.85 \\
    $\quad$Qwen2 VL Instruct 2B & 40.97 & 3.33 & 44.40 & 4.60 & 47.07 & 3.43 & 48.37 & 12.50 & 45.20 & 5.96 \\
    $\quad$Qwen2 VL Instruct 7B & 63.83 & 4.07 & 67.20 & 8.57 & 57.00 & 3.90 & 56.80 & 21.23 & 61.21 & 9.44 \\
    $\quad$Qwen2 VL Instruct 72B & 76.13 & 10.40 & 81.63 & 17.43 & 67.23 & 6.27 & 56.73 & 26.07 & 70.43 & 15.04 \\
    $\quad$Llama 3.2 Instruct 11B & 57.93 & 14.37 & 65.57 & 19.20 & 56.27 & \underline{9.50} & 46.60 & 27.23 & 56.59 & 17.58 \\
    $\quad$Llama 3.2 Instruct 90B & 77.33 & 14.27 & \underline{83.43} & 22.30 & 71.23 & 9.00 & \underline{64.70} & 29.73 & 74.17 & 18.82 \\
    $\quad$Molmo-E 1B & 21.87 & 0.00 & 24.53 & 0.13 & 20.23 & 0.00 & 19.60 & 1.27 & 21.56 & 0.35 \\
    $\quad$Molmo-D 7B & 50.67 & 1.00 & 57.00 & 2.23 & 48.67 & 1.73 & 36.73 & 11.70 & 48.27 & 4.16 \\
    $\quad$Molmo-O 7B & 46.03 & 2.13 & 43.27 & 4.37 & 41.60 & 2.10 & 26.83 & 9.03 & 39.43 & 4.41 \\
    $\quad$Pangea 7B & 45.33 & 0.43 & 59.40 & 1.33 & 22.17 & 0.63 & 34.10 & 17.90 & 40.25 & 5.07 \\
    $\quad$Pangea 7B$^\ddagger$ & 54.87 & 0.43 & 65.77 & 1.33 & 55.00 & 0.63 & 48.47 & 17.90 & 56.03 & 5.07 \\
    $\quad$Aria 25B & 65.77 & 2.67 & 71.43 & 6.47 & 57.13 & 1.80 & 39.60 & 15.70 & 58.48 & 6.66 \\
    $\quad$Phi-3.5 Vision 4B & 49.27 & 1.90 & 53.03 & 3.03 & 42.90 & 1.33 & 31.23 & 8.43 & 44.11 & 3.67 \\
    $\quad$Pixtral 12B & 57.57 & 0.60 & 72.33 & 1.83 & 55.40 & 0.57 & 44.73 & 12.83 & 57.51 & 3.96 \\
    $\quad$NVLM-D 72B & 75.50 & 3.13 & 78.20 & 7.37 & 54.67 & 1.37 & 54.13 & 17.40 & 65.62 & 7.32 \\
    
    \midrule
    \textbf{Proprietary} & & & & & & & & & & \\
    $\quad$GPT-4o & \textbf{88.40} & \textbf{16.60} & \textbf{90.43} & \textbf{35.47} & \textbf{82.23} & \textbf{12.60} & 63.60 & \textbf{35.53} & \textbf{81.17} & \textbf{25.05} \\
    $\quad$GPT-4o Mini & 75.33 & 7.30 & 83.00 & 17.67 & 64.83 & 3.53 & 52.87 & 26.90 & 69.01 & 13.85 \\
    $\quad$Gemini 1.5 Flash & \underline{78.17} & \underline{16.30} & 82.07 & \underline{23.53} & \underline{71.33} & 7.33 & \textbf{66.00} & \underline{32.30} & \underline{74.39} & \underline{19.86} \\
    \midrule

    \textbf{Test Large (60k)} \\ \midrule
    \textbf{Open-Source} & & & & & & & & & & \\ 
    $\quad$Llava1.6 Vicuna 7B & 34.57 & 1.59 & 43.48 & 4.03 & 34.84 & 1.41 & 32.24 & 9.29 & 36.28 & 4.08 \\
    $\quad$Llava1.6 Vicuna 13B & 40.17 & 2.79 & 48.17 & 5.85 & 39.05 & 2.57 & 37.79 & 10.16 & 41.30 & 5.34 \\
    $\quad$Qwen2 VL Instruct 2B & 41.65 & 7.98 & 42.29 & 8.13 & 39.69 & 6.74 & 47.85 & 14.55 & 42.87 & 9.35 \\
    $\quad$Qwen2 VL Instruct 7B & 61.48 & 6.76 & 67.85 & 10.36 & 53.52 & 6.12 & 55.90 & 21.03 & 59.69 & 11.07 \\
    $\quad$Qwen2 VL Instruct 72B & 74.19 & 12.67 & 80.79 & 21.31 & 62.43 & 8.37 & 61.90 & 27.27 & 69.83 & 17.40 \\
    $\quad$Llama 3.2 Instruct 11B & 59.93 & \underline{18.75} & 64.12 & 22.96 & 53.17 & \underline{13.39} & 57.93 & \underline{31.58} & 58.79 & \underline{21.67} \\
    $\quad$Llama 3.2 Instruct 90B  & \underline{77.69} & 16.93 & \underline{82.92} & \underline{23.60} & 63.96 & 10.87 & \underline{67.87} & 31.31 & 73.11 & 20.68 \\
    $\quad$Molmo-E 1B & 18.81 & 0.01 & 24.22 & 0.23 & 19.55 & 0.01 & 18.97 & 1.54 & 20.39 & 0.45 \\
    $\quad$Molmo-D 7B & 46.01 & 2.89 & 55.95 & 3.66 & 41.61 & 2.31 & 33.35 & 11.45 & 44.23 & 5.08 \\
    $\quad$Molmo-O 7B & 39.96 & 5.15 & 44.93 & 6.03 & 38.41 & 3.51 & 29.81 & 10.07 & 38.28 & 6.19 \\
    $\quad$Pangea 7B & 41.38 & 1.52 & 57.95 & 2.73 & 21.77 & 1.57 & 37.15 & 20.15 & 39.56 & 6.49 \\
    $\quad$Pangea 7B$^\ddagger$ & 52.35 & 1.52 & 63.07 & 2.73 & 49.17 & 1.57 & 48.71 & 20.15 & 53.33 & 6.49 \\
    $\quad$Aria 25B & 58.61 & 4.99 & 69.29 & 9.17 & 52.82 & 3.39 & 42.82 & 16.20 & 55.89 & 8.44 \\
    $\quad$Phi-3.5 Vision 4B & 43.37 & 2.91 & 48.71 & 4.23 & 40.87 & 2.07 & 35.01 & 9.22 & 41.99 & 4.61 \\
    $\quad$Pixtral 12B & 56.65 & 1.22 & 70.69 & 2.94 & 52.12 & 1.09 & 46.67 & 14.43 & 56.53 & 4.92 \\
    $\quad$NVLM-D 72B & 69.82 & 4.71 & 78.93 & 10.29 & 52.12 & 2.89 & 51.97 & 16.68 & 63.21 & 8.64 \\
    
    \midrule
    \textbf{Proprietary} & & & & & & & & & & \\ 
    $\quad$GPT-4o & \textbf{88.45} & \textbf{21.88} & \textbf{91.57} & \textbf{37.51} & \textbf{82.29} & \textbf{14.79} & 66.52 & \textbf{37.13} & \textbf{82.21} & \textbf{27.83} \\
    $\quad$GPT-4o Mini & 72.80 & 10.28 & 81.65 & 20.87 & 57.76 & 5.72 & 52.37 & 25.79 & 66.14 & 15.66 \\
    $\quad$Gemini 1.5 Flash & 77.05 & 12.81 & 80.97 & 15.16 & \underline{69.13} & 6.46 & \textbf{71.53} & 30.03 & \underline{74.67} & 16.12 \\

    \bottomrule
    \end{tabular}
}
\caption{Accuracy (\%) results of $\vqaname$. MCQ and OEQ indicate multiple-choice question and open-ended question, respectively. Best and second-best are \textbf{bolded} and \underline{underlined}, respectively. $^\ddagger$We employ an optimized prompt provided by the authors (see Subsection~\ref{prompt-sensitivity} in the Appendix for further details).}
\label{tab:all_results_accuracy}
\end{table*}

\begin{figure*}[!t]
    \centering
    \includegraphics[width=0.95\linewidth]{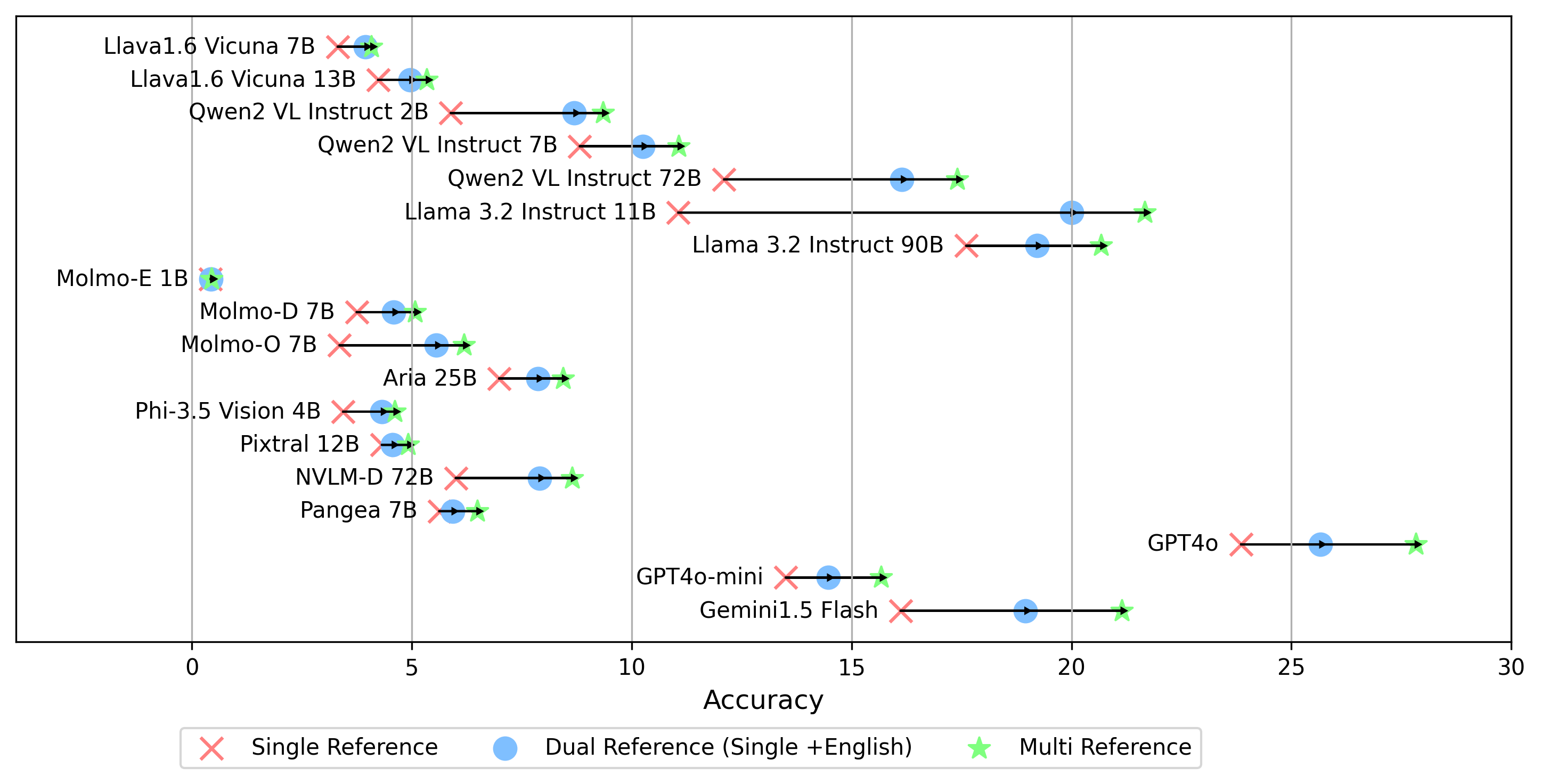}
    \caption{Model performance with different references on open-ended question.}
    \label{fig:models_scatter}
\end{figure*}

\begin{table*}[!th]
\centering
\resizebox{0.93\textwidth}{!}{
    \begin{tabular}{lrrr|r|r}
    \toprule
    \multicolumn{1}{l}{\multirow{2}{*}{\textbf{Model (BERTScore)}}} & \multicolumn{3}{c|}{\textbf{Task 1 (Dish Name)}} & \multicolumn{1}{c|}{\multirow{2}{*}{\textbf{\makecell{Task 2 \\ (Location)}}}} & \multicolumn{1}{c}{\multirow{2}{*}{\textbf{Average}}} \\ 
    & \multicolumn{1}{c}{\textbf{(a) no-context}} & \multicolumn{1}{c}{\textbf{(b) contextualized}} & \multicolumn{1}{c|}{\textbf{(c) adversarial}} & & \\
    \midrule
    \textbf{Test Small (12k)} \\ \midrule
    \textbf{Open-Source} & & & & & \\
    $\quad$Llava1.6 Vicuna 7B & 81.49 & 82.13 & 81.56 & 85.45 & 82.66 \\
    $\quad$Llava1.6 Vicuna 13B & 80.50 & 80.65 & 80.14 & 81.77 & 80.77 \\
    $\quad$Qwen2 VL Instruct 2B & 82.48 & 82.75 & 82.34 & 84.29 & 82.97 \\
    $\quad$Qwen2 VL Instruct 7B & 82.65 & 83.13 & 82.10 & 87.22 & 83.78 \\
    $\quad$Qwen2 VL Instruct 72B & 83.78 & 84.63 & 83.06 & 87.10 & 84.64 \\
    $\quad$Llama 3.2 Instruct 11B & 82.45 & 82.93 & 81.64 & 82.59 & 82.40 \\
    $\quad$Llama 3.2 Instruct 90B & 82.82 & 83.44 & 81.98 & 85.70 & 83.48 \\
    $\quad$Molmo-E 1B & 81.17 & 81.12 & 81.24 & 83.58 & 81.78 \\
    $\quad$Molmo-D 7B & 81.26 & 81.65 & 80.55 & 84.87 & 82.08 \\
    $\quad$Molmo-O 7B & 82.14 & 82.24 & 81.44 & 84.38 & 82.55 \\
    $\quad$Pangea 7B & 81.29 & 81.78 & 80.19 & 86.31 & 82.39 \\
    $\quad$Aria 25B & 79.85 & 80.26 & 79.86 & 80.53 & 80.12 \\
    $\quad$Phi-3.5 Vision 4B & 80.82 & 79.66 & 76.77 & 83.25 & 80.12 \\
    $\quad$Pixtral 12B & 78.84 & 79.12 & 78.90 & 86.40 & 80.81 \\
    $\quad$NVLM-D 72B & 81.39 & 82.05 & 79.98 & 85.64 & 82.27 \\
    \midrule
    \textbf{Proprietary} & & & & & \\
    $\quad$GPT-4o & \textbf{84.86} & \textbf{86.92} & \textbf{83.89} & \underline{88.98} & \textbf{86.16} \\
    $\quad$GPT-4o Mini & 83.10 & 83.91 & 82.16 & 87.34 & 84.13 \\
    $\quad$Gemini 1.5 Flash & \underline{84.68} & \underline{85.09} & \underline{83.11} & \textbf{89.15} & \underline{85.51} \\
    \midrule
    \textbf{Test Large (60k)} \\ \midrule
    \textbf{Open-Source} & & & & & \\
    $\quad$Llava1.6 Vicuna 7B & 81.63 & 82.10 & 81.58 & 85.81 & 82.78 \\
    $\quad$Llava1.6 Vicuna 13B & 80.65 & 80.70 & 80.12 & 81.86 & 80.83 \\
    $\quad$Qwen2 VL Instruct 2B & 82.95 & 83.10 & 82.81 & 84.51 & 83.34 \\
    $\quad$Qwen2 VL Instruct 7B & 82.92 & 83.42 & 82.30 & 87.39 & 84.01 \\
    $\quad$Qwen2 VL Instruct 72B & 83.72 & \underline{85.10} & 83.11 & 87.42 & 84.84 \\
    $\quad$Llama 3.2 Instruct 11B & 82.54 & 82.79 & 81.64 & 82.88 & 82.46 \\
    $\quad$Llama 3.2 Instruct 90B & 83.05 & 83.51 & 81.95 & 85.85 & 83.59 \\
    $\quad$Molmo-E 1B & 81.17 & 81.10 & 81.13 & 83.87 & 81.82 \\
    $\quad$Molmo-D 7B & 81.39 & 81.63 & 80.73 & 85.10 & 82.21 \\
    $\quad$Molmo-O 7B & 82.27 & 82.21 & 81.52 & 84.63 & 82.66 \\
    $\quad$Pangea 7B & 81.40 & 81.91 & 80.23 & 86.79 & 82.58 \\
    $\quad$Aria 25B & 79.89 & 80.20 & 79.83 & 80.63 & 80.14 \\
    $\quad$Phi-3.5 Vision 4B & 80.98 & 79.55 & 77.61 & 83.31 & 80.36 \\
    $\quad$Pixtral 12B & 79.00 & 79.33 & 78.98 & 86.75 & 81.02 \\
    $\quad$NVLM-D 72B & 81.54 & 82.17 & 80.05 & 85.67 & 82.36 \\
    \midrule
    \textbf{Proprietary} & & & & & \\
    $\quad$GPT-4o & \textbf{85.04} & \textbf{86.93} & \textbf{83.92} & \underline{89.06} & \textbf{86.24} \\
    $\quad$GPT-4o Mini & 83.19 & 84.05 & 82.38 & 87.30 & 84.23 \\
    $\quad$Gemini 1.5 Flash & \underline{84.47} & 84.97 & \underline{83.14} & \textbf{89.43} & \underline{85.50} \\
    \bottomrule
    \end{tabular}
}
\caption{BERTScore results of $\vqaname$. Only the results from open-ended (OEQ) are used. Best and second-best are \textbf{bolded} and \underline{underlined}, respectively.}
\label{tab:all_results_bertscore}
\end{table*}

\paragraph{Robustness and Error Analysis.}
Figure~\ref{fig:bert_score_scatter} illustrates the correlation between BERTScore and accuracy in the open-ended setting through regression analysis. The R-squared value is 0.41, indicating a low correlation between BERTScore and accuracy. Despite this, BERTScore remains a useful metric for assessing whether the model's predictions have semantic similarity to the gold labels, even if they are not exact matches.

\section{Evaluation}
\subsection{Prompt Sensitivity}
\label{prompt-sensitivity}
We use the same prompts for all models, with the exception of the Pangea 7B model~\cite{yue2024pangea}. This model is particularly sensitive and lacks robustness in handling diverse prompt instructions, often struggling to follow instructions accurately, especially in multiple-choice questions (MCQs), unless a specific template is applied. In contrast, models like Llama 3.2 Instruct and Qwen2 VL Instruct are more adaptable to varied instructions. After consulting with the authors, we adopted the prompt ``Answer with the option letter from the given choices directly.'' for MCQ queries when using the Pangea 7B model.

\begin{table*}[!th]
\centering
\resizebox{0.65\textwidth}{!}{
    \begin{tabular}{lrrr}
    \toprule
    \textbf{Continents/Regions} & \multicolumn{1}{c}{\textbf{\# Countries}} & \multicolumn{1}{c}{\textbf{\# Food Entries}} & \multicolumn{1}{c}{\textbf{\% in Our Data}} \\ \midrule
    \textbf{Global}$^*$ & N/A$\quad$ & \textbf{96}$\quad$ & \textbf{3.98\%}$\quad$ \\
    \textbf{Africa} & \textbf{52}$\quad$ & \textbf{190}$\quad$ & \textbf{7.87\%}$\quad$ \\
        $\quad$Eastern Africa & 18 & 40 & 1.7\% \\
        $\quad$Middle Africa & 6 & 17 & 0.7\% \\
        $\quad$Northern Africa & 7 & 67 & 2.8\% \\
        $\quad$Southern Africa & 5 & 33 & 1.4\% \\
        $\quad$Western Africa & 16 & 60 & 2.5\% \\
    \textbf{America} & \textbf{37}$\quad$ & \textbf{472}$\quad$ & \textbf{19.55\%}$\quad$ \\
        $\quad$Caribbean & 15 & 60 & 2.5\% \\
        $\quad$Central America & 8 & 134 & 5.6\% \\
        $\quad$Northern America & 2 & 230 & 9.5\% \\
        $\quad$South America & 12 & 109 & 4.5\% \\
    \textbf{Europe} & \textbf{47}$\quad$ & \textbf{808}$\quad$ & \textbf{33.47\%}$\quad$ \\
        $\quad$Eastern Europe & 10 & 164 & 6.8\% \\
        $\quad$Northern Europe & 15 & 237 & 9.8\% \\
        $\quad$Southern Europe & 13 & 300 & 12.4\% \\
        $\quad$Western Europe & 9 & 233 & 9.7\% \\
    \textbf{Asia} & \textbf{53}$\quad$ & \textbf{1,052}$\quad$ & \textbf{43.58\%}$\quad$ \\
        $\quad$Central Asia & 5 & 10 & 0.4\% \\
        $\quad$Eastern Asia & 9 & 420 & 17.4\% \\
        $\quad$South Eastern Asia & 12 & 362 & 15.0\% \\
        $\quad$Southern Asia & 9 & 200 & 8.3\% \\
        $\quad$Western Asia & 18 & 155 & 6.4\% \\
    \textbf{Oceania} & \textbf{3}$\quad$ & \textbf{37}$\quad$ & \textbf{1.53\%}$\quad$  \\
        $\quad$Australia \& New Zealand & 2 & 33 & 1.4\% \\
        $\quad$Melanesia & 1 & 4 & 0.2\% \\
        $\quad$Micronesia & - & - & - \\
        $\quad$Polynesia & - & - & - \\
    \bottomrule
    \end{tabular}
}
\caption{
    Geographical distribution of \kbname, corresponds to Figure \ref{fig:dish-geodist}. Note that there are food entries linked to multiple regions, with some linked to multiple continents. $^*$\textbf{Global} denotes entries with more than five regions.
}
\label{tab:region_dist}
\end{table*}

\begin{figure*}[!th]
    \centering
    \includegraphics[width=\linewidth]{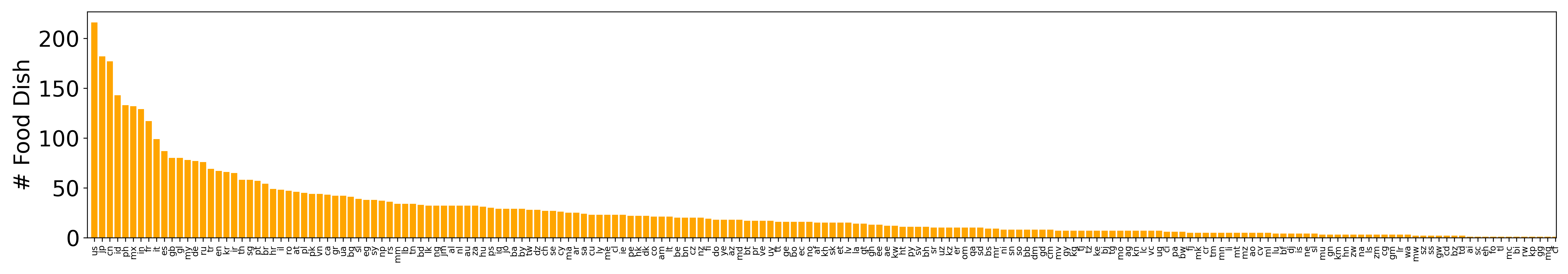} 
    \caption{Dish frequency by country showing 189 countries.}
    \label{fig:all_country_food_dish}
\end{figure*}

\begin{table*}[!th]
    \centering
    \resizebox{\linewidth}{!}{
    \begin{tabular}{lcrr}
        \toprule
        \textbf{Country} & \textbf{Count} & \textbf{\%} \\
        \midrule
        United States & 216 & 9.47\\
        Japan & 182 & 7.98\\
        China & 177 & 7.76\\
        Indonesia & 143 & 6.27\\
        Philippines & 133 & 5.83\\
        Mexico & 132 & 5.78\\
        India & 129 & 5.65\\
        France & 117 & 5.12\\
        Italy & 99 & 4.34\\
        Spain & 87 & 3.81\\
        United Kingdom & 80 & 3.51\\
        Global & 80 & 3.51\\
        Germany & 77 & 3.37\\
        Russia & 76 & 3.33\\
        Turkey & 69 & 3.02\\
        Korea & 66 & 2.89\\
        Iran & 65 & 2.85\\
        Thailand & 58 & 2.54\\
        Singapore & 58 & 2.54\\
        Portugal & 57 & 2.50\\
        Brazil & 54 & 2.37\\
        Israel & 48 & 2.10\\
        Romania & 47 & 2.06\\
        Austria & 46 & 2.02\\
        Poland & 45 & 1.97\\
        Pakistan & 44 & 1.93\\
        Vietnam & 44 & 1.93\\
        Canada & 43 & 1.89\\
        Greece & 42 & 1.84\\
        Ukraine & 42 & 1.84\\
        Bulgaria & 41 & 1.80\\
        Slovenia & 39 & 1.71\\
        Egypt & 38 & 1.67\\
        Syria & 38 & 1.67\\
        Nepal & 37 & 1.62\\
        Serbia & 36 & 1.58\\
        Myanmar & 34 & 1.49\\
        Lebanon & 34 & 1.49\\
        Tunisia & 34 & 1.49\\
        Bangladesh & 33 & 1.45\\
        Malaysia & 32 & 1.40\\
        Sri Lanka & 32 & 1.40\\
        Nigeria & 32 & 1.40\\
        Jamaica & 32 & 1.40\\
        Netherlands & 32 & 1.40\\
        Albania & 32 & 1.40\\
        South Africa & 32 & 1.40\\
        Australia & 32 & 1.40\\
        Hungary & 31 & 1.36\\
        Palestine & 30 & 1.32\\
        Iraq & 29 & 1.27\\
        Jordan & 29 & 1.27\\
        Bosnia and Herzegovina & 29 & 1.27\\
        Taiwan & 28 & 1.23\\
        Algeria & 28 & 1.23\\
        Switzerland & 27 & 1.18\\
        Cyprus & 26 & 1.14\\
        Morocco & 25 & 1.10\\ \bottomrule
    \end{tabular}
    \begin{tabular}{lcrr}
        \toprule
        \textbf{Country} & \textbf{Count} & \textbf{\%} \\
        \midrule
        Argentina & 25 & 1.10\\
        Saudi Arabia & 24 & 1.05\\
        North Macedonia & 24 & 1.05\\
        Cuba & 23 & 1.01\\
        Libya & 23 & 1.01\\
        Montenegro & 23 & 1.01\\
        Chile & 23 & 1.01\\
        Ireland & 23 & 1.01\\
        Peru & 22 & 0.96\\
        Hong Kong & 22 & 0.96\\
        Denmark & 22 & 0.96\\
        Colombia & 21 & 0.92\\
        Armenia & 21 & 0.92\\
        Lithuania & 21 & 0.92\\
        Belgium & 20 & 0.88\\
        Brunei Darussalam & 20 & 0.88\\
        Czech Republic & 20 & 0.88\\
        New Zealand & 20 & 0.88\\
        Finland & 19 & 0.83\\
        Dominican Republic & 18 & 0.79\\
        Yemen & 18 & 0.79\\
        Azerbaijan & 18 & 0.79\\
        Moldova & 18 & 0.79\\
        Bhutan & 17 & 0.75\\
        Puerto Rico & 17 & 0.75\\
        Venezuela & 17 & 0.75\\
        Uruguay & 17 & 0.75\\
        Bolivia & 16 & 0.70\\
        Trinidad and Tobago & 16 & 0.70\\
        Georgia & 16 & 0.70\\
        Norway & 16 & 0.70\\
        Cambodia & 15 & 0.66\\
        Afghanistan & 15 & 0.66\\
        Slovakia & 15 & 0.66\\
        Ethiopia & 15 & 0.66\\
        Latvia & 15 & 0.66\\
        Laos & 14 & 0.61\\
        Guatemala & 14 & 0.61\\
        Ghana & 13 & 0.57\\
        United Arab Emirates & 12 & 0.53\\
        Kuwait & 12 & 0.53\\
        Paraguay & 11 & 0.48\\
        El Salvador & 11 & 0.48\\
        Bahrain & 11 & 0.48\\
        Haiti & 11 & 0.48\\
        Uzbekistan & 10 & 0.44\\
        Kazakhstan & 10 & 0.44\\
        Eritrea & 10 & 0.44\\
        Oman & 10 & 0.44\\
        Qatar & 10 & 0.44\\
        Sudan & 10 & 0.44\\
        Suriname & 10 & 0.44\\
        Mauritania & 9 & 0.39\\
        Bahamas & 9 & 0.39\\
        Nicaragua & 8 & 0.35\\
        Senegal & 8 & 0.35\\
        Barbados & 8 & 0.35\\ 
        Dominica & 8 & 0.35\\ \bottomrule
    \end{tabular}
    \begin{tabular}{lcrr}
        \toprule
        \textbf{Country} & \textbf{Count} & \textbf{\%} \\
        \midrule
        Grenada & 8 & 0.35\\
        Cameroon & 8 & 0.35\\
        Somalia & 8 & 0.35\\
        Antigua and Barbuda & 7 & 0.31\\
        Maldives & 7 & 0.31\\
        Kyrgyzstan & 7 & 0.31\\
        Tajikistan & 7 & 0.31\\
        Togo & 7 & 0.31\\
        Uganda & 7 & 0.31\\
        Benin & 7 & 0.31\\
        Macau & 7 & 0.31\\
        Guyana & 7 & 0.31\\
        Saint Kitts and Nevis & 7 & 0.31\\
        Saint Lucia & 7 & 0.31\\
        Saint Vincent and the Grenadines & 7 & 0.31\\
        Fiji & 5 & 0.22\\
        Mongolia & 5 & 0.22\\
        Liechtenstein & 5 & 0.22\\
        Macedonia & 5 & 0.22\\
        Malta & 5 & 0.22\\
        Mozambique & 5 & 0.22\\
        Angola & 5 & 0.22\\
        Cabo Verde & 5 & 0.22\\
        Turkmenistan & 5 & 0.22\\
        Costa Rica & 5 & 0.22\\
        Burkina Faso & 4 & 0.18\\
        Luxembourg & 4 & 0.18\\
        Djibouti & 4 & 0.18\\
        Iceland & 4 & 0.18\\
        Sierra Leone & 4 & 0.18\\
        Niger & 4 & 0.18\\
        Mauritius & 3 & 0.13\\
        Guinea & 3 & 0.13\\
        Zimbabwe & 3 & 0.13\\
        Namibia & 3 & 0.13\\
        Lesotho & 3 & 0.13\\
        Zambia & 3 & 0.13\\
        Congo & 3 & 0.13\\
        Gambia & 3 & 0.13\\
        Liberia & 3 & 0.13\\
        Comoros & 3 & 0.13\\
        South Korea & 3 & 0.13\\
        Wales & 3 & 0.13\\
        Honduras & 3 & 0.13\\
        Anguilla & 1 & 0.04\\
        Western Sahara & 1 & 0.04\\
        Faroe Islands & 1 & 0.04\\
        Seychelles & 1 & 0.04\\
        Burundi & 1 & 0.04\\
        Rwanda & 1 & 0.04\\
        North Korea & 1 & 0.04\\
        Timor-Leste & 1 & 0.04\\
        Guernsey & 1 & 0.04\\
        Madagascar & 1 & 0.04\\
        Central African Republic & 1 & 0.04\\
        Monaco & 1 & 0.04 \\
        \bottomrule \\
        \\
    \end{tabular}
    }
    \caption{Distribution of food entries by country.}
    \label{tab:food_entries}
\end{table*}

\end{document}